\documentclass[11pt, a4paper, onecolumn, copyright, goog, article]{google}

\usepackage[authoryear, sort&compress, round]{natbib}
\usepackage{float}

\usepackage[capitalize]{cleveref}
\usepackage{multicol}

\usepackage{xcolor}
\usepackage{tcolorbox}

\usepackage[utf8]{inputenc}
\usepackage{booktabs}
\usepackage{multirow}

\definecolor{AgentBlue}{HTML}{0056B3} 
\definecolor{EnvOchre}{HTML}{2E7D32}
\definecolor{ProbeGrey}{HTML}{555555}

\usepackage[scaled=0.9]{beramono}
\usepackage{array}
\newcommand{\scaffold}[1]{{\color{EnvOchre}\ttfamily\noindent #1}}

\newcommand{\agent}[1]{{\color{AgentBlue}\texttt{#1}}}

\definecolor{PaleYellow}{HTML}{FFFEF2}

\uselogo{} 

\title{Stabilising Generative Models of Attitude Change}

\correspondingauthor{jmatyas@deepmind.com}

\author[1]{Jayd Matyas}
\author[1,2]{William A. Cunningham}
\author[1]{Alexander Sasha Vezhnevets}
\author[3]{Dean Mobbs}
\author[1,$\dagger$]{Edgar A. Duéñez-Guzmán}
\author[1]{Joel Z. Leibo}

\affil[1]{\thepa{}{}}
\affil[2]{University of Toronto}
\affil[3]{Caltech}
\affil[$\dagger$]{\textit{Currently at Gibran AI}}

\begin{abstract}
 
Attitude change---the process by which individuals revise their evaluative stances---has been explained by a set of influential but competing verbal theories. These accounts often function as mechanism sketches: rich in conceptual detail, yet lacking the technical specifications and operational constraints required to run as executable systems. We present a generative actor-based modelling workflow for `rendering' these sketches as runnable actor--environment simulations using the Concordia simulation library \citep{vezhnevets2023generative, vezhnevets2025multi}. In Concordia, actors operate by predictive pattern completion: an operation on natural language strings that generates a suffix which describes the actor's intended action from a prefix containing memories of their past and observations of the present. We render the theories of cognitive dissonance \citep{festinger1957cognitive}, self-consistency \citep{aronson1969theory}, and self-perception \citep{bem1972self} as distinct decision logics that populate and process the prefix through theory-specific sequences of reasoning steps. We evaluate these implementations across classic psychological paradigms, operationalising them through their canonical experimental proxies: choice difficulty (item rating experiment; \citealp{brehm1956postdecision}), forced compliance (boring task experiment; \citealp{festinger1959cognitive}), and anticipatory suffering (worm experiment; \citealp{comer1975choosing}). We further evaluate these benchmarks alongside self-affirmation manipulations \citep{steele1988psychology}. Our implementations generate behavioural patterns consistent with known results from the original empirical literature. However, we find that achieving stable reproduction requires resolving the inherent underdetermination of the verbal accounts and the conflicts between modern linguistic priors and historical experimental assumptions. We document how this manual process of iterative model `stabilisation' surfaces specific operational and socio-ecological dependencies that were largely undocumented in the original verbal accounts. Ultimately, we argue that the manual stabilisation process itself should be regarded as a core part of the methodology, functioning to clarify situational and representational commitments needed to generate characteristic effects.
\end{abstract}

\begin{document}

\maketitle

\section{Introduction}

Attitude change---the process by which an individual’s evaluative stance shifts over time---is among the oldest and most extensively studied phenomena in psychology. Dating back to the 1950s, classic accounts offer sharply different explanations: Cognitive dissonance theory \citep{festinger1957cognitive} proposes that conflict among cognitions generates psychological discomfort, which motivates updating cognitions to restore consistency. Self-consistency theory \citep{aronson1969theory} restricts this conflict to threats against `firm expectations', especially concerning the moral or competent self. Self-perception theory \citep{bem1972self} rejects the need for internal conflict altogether, proposing that individuals infer their own attitudes from observed behaviour in context. A contemporary extension of this view suggests that such rationalisation is actually a computationally rational strategy for belief updating \citep{cushman2020rationalization}. Importantly, these accounts often generate indistinguishable behavioural predictions in canonical paradigmatic experiments. This empirical overlap highlights the importance of contemporary syntheses which emphasise that attitude change effects are highly sensitive to cultural and situational context \citep{albarracin2018attitudes}. Because the mechanisms are so context dependent, specifying the precise conditions under which they operate is essential for disentangling their competing explanations \citep{fazio1977dissonance}.

Psychological mechanisms have historically been explored through two different methodological lenses. Verbal theories, or ‘word-models’, capture conceptual richness but often remain theoretically underdetermined \citep{oberauer2019addressing}; they function as mechanism sketches \citep{fried2020theories} that lack the formal constraints required to be reproduced systematically. Computational models, by contrast, offer rigorous replicability, but typically achieve this by abstracting away the representational and situational details that capture the original psychological phenomena. Recent advances in large language models (LLMs) afford a new space for resolving this logical opacity \citep{guest2021computational}. By using LLM-based generative actors \citep{vezhnevets2023generative, vezhnevets2025multi} embedded within structured simulation environments (similar to the approach of \citealp{park2023generative}), we can construct \textit{in silico} models that explicitly instantiate the mechanistic commitments of a verbal theory using the natural language in which it was originally described.

\begin{figure}[H]
    \centering
    \includegraphics[width=0.85\linewidth]{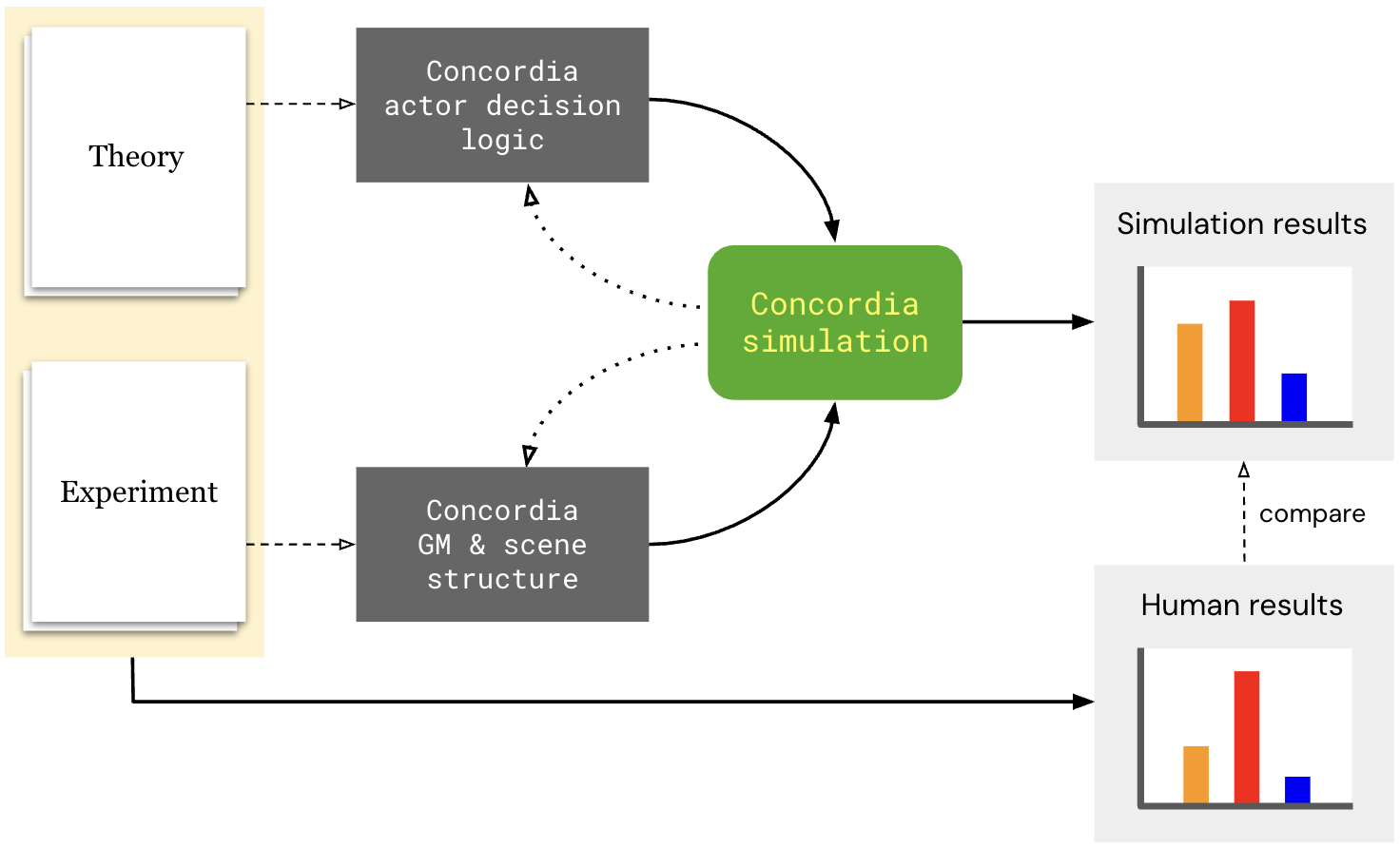}
    \captionsetup{font=scriptsize}
    \caption{The generative actor-based modelling (GABM) rendering and stabilisation workflow. Dashed arrows represent descriptive or representational mappings, solid lines represent direct causal or functional relationships, and dotted lines denote the iterative stabilisation process where simulation outcomes are used to refine the system.}
    \label{fig:flow}
\end{figure}

In the Concordia framework \citep{vezhnevets2023generative, vezhnevets2025multi}, generative actors operate via predictive pattern completion: an operation on natural language strings that generates a suffix describing the actor's intended action from a prefix containing memories of their past and observations of the present. This architecture is situated within discrete scenes moderated by a central Game Master (GM), a structure inspired by tabletop role-playing games such as Dungeons \& Dragons \citep{gygax1974dungeons}. In these games, participants use verbal descriptions to take actions within a shared story, while a central GM adjudicates the results and maintains the narrative world. Within this architecture, the actor’s internal decision logic functions as the theory being tested, while discrete scenes and a GM provide the experimental environment. Building on the finding that LLMs are naturally adept at adopting personas and engaging in role-play \citep{shanahan2023role}, our methodology leverages this capacity to create a formal simulation pipeline.

Using Concordia, we formalise the three foundational accounts of attitude change described above---Festinger's theory of cognitive dissonance, Aronson's theory of self-consistency, and Bem's theory of self-perception---as distinct \textit{decision logics} that operationalise these theoretical mechanisms. Each logic processes the situational prefix through theory-specific sequences of reasoning \citep{leibo2026rationality}. Along with a Minimal baseline (a `naked' LLM, i.e., no decision logic) actor. 

We subject these actors to three canonical paradigmatic experiments from the literature: the Item Rating experiment (choice difficulty paradigm; \citealp{brehm1956postdecision}), the Boring Task experiment (forced compliance paradigm; \citealp{festinger1959cognitive}), and the Worm experiment (anticipatory suffering paradigm; \citealp{comer1975choosing}).

Additionally, we implement a self-affirmation manipulation \citep{steele1988psychology} across all environments. In human subjects, providing an opportunity to affirm a core personal value significantly attenuates or entirely eliminates the attitude change typically induced by these experimental paradigms \citep{steele1983dissonance}.

We show that all three theory-driven actors: Festinger, Aronson, and Bem, reproduce the canonical behavioural shifts in each of the three experiments and in the self-affirmation manipulation version of each\footnote{Because Aronson's theory restricts dissonance to threats against firm expectations such as the moral or competent self, a pre-stimulus intervention that buffers the self-concept should selectively reduce attitude change for this actor. The others fit the self-affirmation manipulation data when the self-generated affirmation happens to be consonant with the task at hand (see Section \ref{sec:self_affirmation_results}).}. Importantly, these behavioural patterns are consistently absent in the Minimal actor, which fails to produce any attitude change across any experimental condition.

In this work, we describe a generative actor-based methodology for rendering verbal psychological theories into executable systems. We show how achieving stable reproduction of effects in the historical psychology literature can require careful manual tuning of the simulation parameters across all actor decision logics and scenarios. This iterative process of stabilising the actor--environment coupling involves manually refining the environmental scaffolding to prevent competing interpretive frames from dominating actor behaviour, thereby surfacing specific semantic and normative scaffolds required for psychological theories to operate that may have been left undocumented in their original accounts. 

Rather than viewing this iterative stabilisation as mere debugging, we consider it a core component of the methodology in its own right. As such, there are better and worse ways to perform iterative stabilisation. By describing our workflow in some detail, we hope to shed light on a part of the modelling process that is not often discussed and provide a positive exemplar for future research. Ultimately, this approach formalises the structural conditions under which observations characteristic of putative psychological mechanisms emerge in our models.

By applying this approach to the study of attitude change, we demonstrate a novel generative actor-based workflow for psychological modelling. While traditional computational approaches often achieve rigour by abstracting away the complexity of the phenomenon itself and its context, our approach instantiates these theories utilising the natural language in which they were originally described. Rather than viewing these classic theories as competing mechanism sketches, we render them as distinct algorithmic processes that can be rigorously isolated and tested under controlled conditions. This approach allows us to disentangle theories that otherwise produce overlapping behavioural predictions, providing a formalised account of the specific situational and representational commitments that generate each account's characteristic effects.

The remainder of the paper proceeds as follows. In Section 2, we introduce our experimental setup using the Concordia framework, detailing how we construct the simulated scenes and generate diverse actor personas. We then formalise the classic verbal theories of Festinger, Aronson, and Bem into executable decision logics in Section 3, with comprehensive details and example traces provided in the Appendices. Section 4 outlines the implementation of our chosen paradigmatic experiments. In Section 5, we present our simulation results, demonstrating that our generative models successfully reproduce the characteristic behavioural patterns found in the original empirical literature. Importantly, Section 5.2 documents the iterative stabilisation process required to achieve these results, revealing hidden contextual and operational dependencies of the original theories. Finally, we conclude with a discussion of the broader implications of this generative workflow for psychological modelling in Section 6.

\section{Experimental set-up: Concordia}
Simulations were built and orchestrated using the Concordia framework, which utilises a structured, scene-based architecture. This structure allows us to embed the necessary constraints of a laboratory experiment directly into the progression of the simulation. Within our simulations, all LLM operations were powered by Gemma 2 \citep{team2024gemma}.

\subsection{Scene structure and the Game Master}

The experimental framework is orchestrated by a central GM that manages the coupling between generative actors and their environment. Within the Concordia framework, the GM functions as the primary engine of the simulated world, responsible for maintaining world state and translating the natural language action descriptions provided by actors into appropriate environmental outcomes. This ensures that while actors are free to describe their intentions in open-ended text, the resulting state changes remain grounded within the strict procedural boundaries of the experimental design.

Concordia utilises a discrete, scene-based architecture. Each scene acts as a temporal container that defines the physical location, the available non-player characters, and any specific events to be executed for a given phase of the experiment. By partitioning the simulation into these discrete stages, the environment allows us to impose the interactional constraints necessary to observe theoretical mechanisms in isolation. This structure enables a controlled interaction where the GM enforces fidelity to the original laboratory protocols while maintaining a coherent narrative context for the actors.

\subsection{Character generation and context setting} \label{sec:character_gen}
To ensure that the observed effects are robust to individual differences, we follow recent advances in population synthesis for generative agents \citep{argyle2023out}. Each actor is initialised with a unique persona anchored by a 16-year generational cohort and a randomised Big Five personality profile. This initialisation aims to establish the psychometric consistency required for valid personality-driven behaviour \citep{serapio2023personality}. These traits are then used to generate and seed a series of formative background memories, providing the memory content required for the actors to act as coherent social actors throughout the simulation. The Big Five scores were only used to condition the formative memory generator and were discarded afterwards.

This context setting extends beyond general biography to establish a narrative bridge to the laboratory setting: each actor generates a unique memory detailing how they specifically encountered the call for participants (for example, through a university bulletin board or a professional advertisement), with the specific medium of discovery determined by the actor's unique character and history.

Furthermore, we found that maintaining the actor’s psychological stability required providing a contextually appropriate pre-experiment location. In early iterations, we hardcoded a shared starting location (such as a local café) for all actors immediately prior to their lab appointment. However, this often induced unintended dissonance (observed in our dissonance-based models), as a public social setting frequently conflicted with the specific needs or personalities of certain actors. Stabilising the simulation required allowing each actor to generate a unique pre-experimental environment (ranging from their own homes to specific workplaces) that aligned with their character, ensuring that their entry into the laboratory was grounded in a coherent and stable situational context.

These design choices function as parameters shaping the resulting actor--environment dynamics; robustness considerations for these choices are discussed in \cref{sec:discussion}.

\begin{figure}[t]
    \centering
    \includegraphics[width=0.85\linewidth]{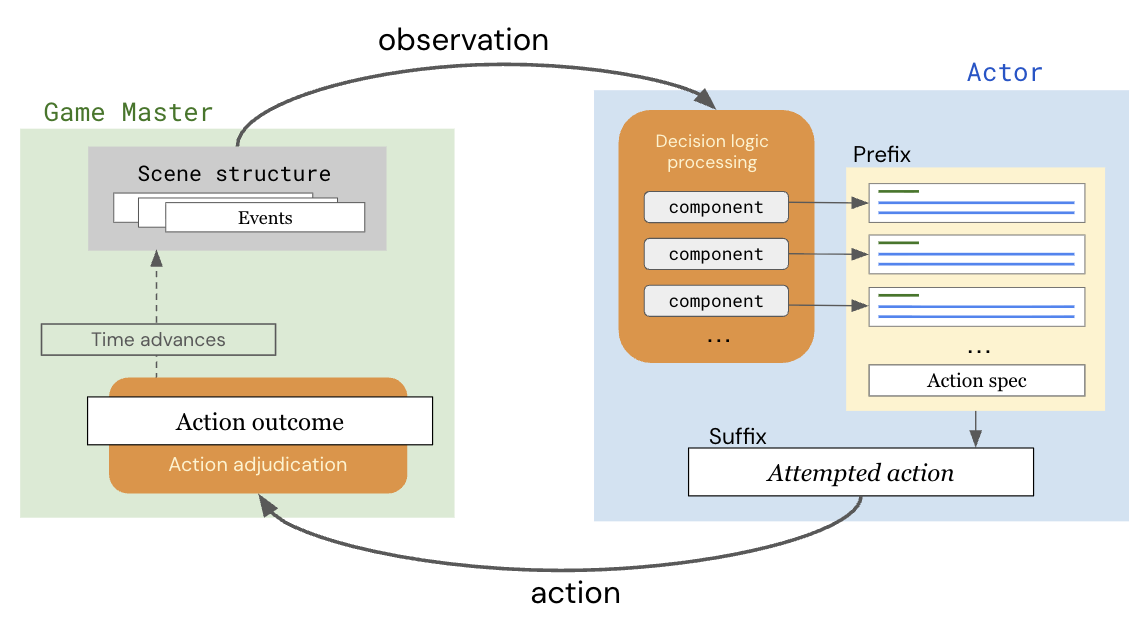}
    \captionsetup{font=scriptsize}
    \caption{The generative actor--environment interaction cycle. The actor processes observations and memories through a theory-specific decision logic made up of various interacting components to form a prefix. The actor completes this pattern with a suffix (the attempted action), which the GM then adjudicates to advance the world state and time.}
    \label{fig:interaction_pattern}
\end{figure}

\section{Psychological theories as Concordia decision logics}

To translate verbal psychological theories into executable computational systems, we must formalise the internal cognitive processes that bridge an actor's environmental observations and their subsequent actions. Within the Concordia framework, this individual decision-making process is modelled via predictive pattern completion \citep{vezhnevets2023generative, leibo2024theory, leibo2026rationality}.

Rather than generating an action in a single unconstrained LLM prediction step, actor cognition is structured in Concordia by a sequence of intermediate computational steps. We refer to this structured sequence as a \emph{decision logic}.

A decision logic is interpreted as an algorithmic-level description (following \citealp{marr1976understanding}) of how information is organised and transformed within the actor's mind. Computationally, it is implemented by chaining together distinct steps, functioning as a series of self-posed questions \citep{leibo2026rationality}. The output of one step becomes part of the context window for the next, iteratively shaping the contents of the prefix—the text-based prompt provided to the actor’s LLM. From this prefix, the model generates a suffix: a string of natural language text describing the actor's intended action (see \autoref{fig:interaction_pattern}). This suffix is then passed to the GM, who adjudicates the action's outcome within the simulated world, updates the environment, and advances the simulation's clock to the next timestep. 

We treat each verbally expressed psychological theory as a distinct decision logic, converting their abstract mechanism sketches into concrete algorithmic procedures. The core differences between theories of attitude change are therefore operationalised as differences in the specific chain of queries an actor uses to process its observations before generating its choice of action.

{\sloppy
\subsection{Decision logics of Festinger, Aronson, and Bem actors} \label{sec:actor_logics}

In this section, we treat each account of attitude change as a design constraint on actor decision logics. As each verbal theory lacks the technical details necessary for computational implementation and is effectively an ablation of a larger cognitive structure, we must interpret the original texts to inform concrete operational decisions. This process necessitates navigating long-standing theoretical ambiguities: a critical question for modelling, for instance, is whether attitudes should be treated as stable, pre-stored representations \citep{fazio1977dissonance} or as evaluations constructed on the spot from available memory \citep{schwarz2007attitude}. As illustrated in \autoref{fig:components}, these decision logics are implemented by chaining together distinct components that transform situational and representational details within the actor’s internal decision logic. Our goal is to document the representational choices we found sufficient to enable each account to run as an executable system within shared paradigmatic environments and behave as expected.

\subsubsection{Dissonance-based decision logics (Festinger \& Aronson actors)}
These actors share a common logic for synthesising observations into cognitions. These components serve a dual purpose: they populate the actor’s prefix to ground the current simulation step and provide the necessary data for evaluating internal inconsistency.

\begin{figure}[b]
    \centering
    \includegraphics[width=0.95\linewidth]{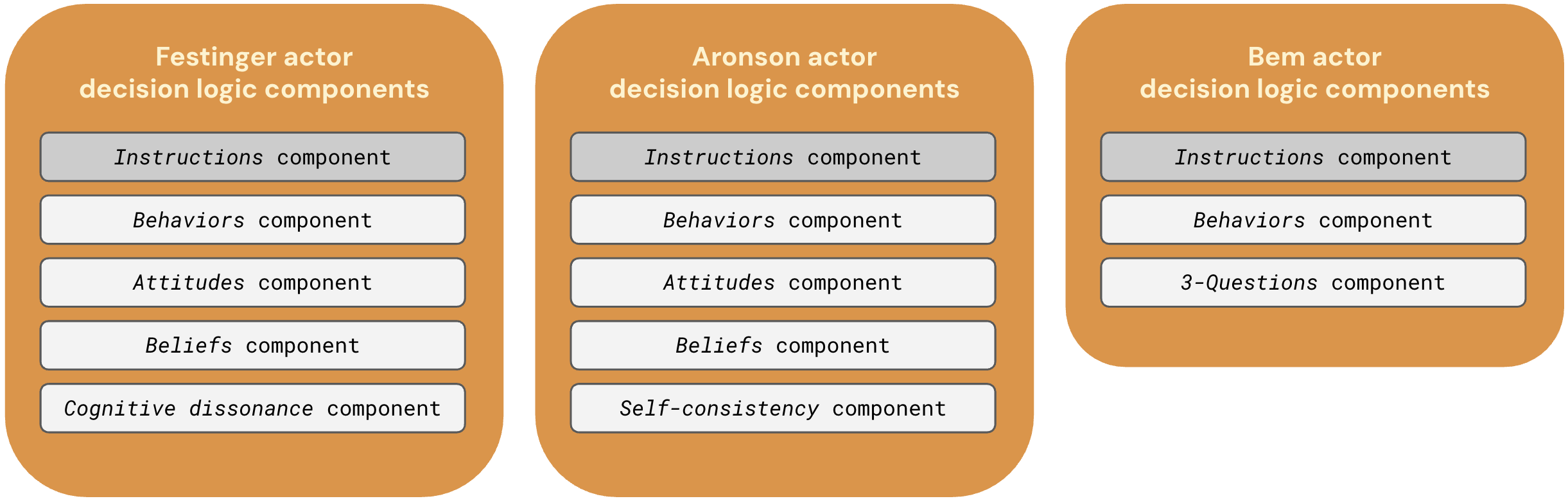}
    \captionsetup{font=scriptsize}
    \caption{Architectural comparison of theory-driven actor decision logics. While all actors share a fixed \textit{Instructions} component and a \textit{Behaviors} component, the Festinger and Aronson actors additionally share \textit{Attitudes} and \textit{Beliefs} components before diverging at their respective conflict evaluation mechanisms. In contrast, the Bem actor utilises a parsimonious sequence of three self-reflexive queries.}
    \label{fig:components}
\end{figure}

\paragraph{\textit{Behaviors} component}

This component produces a situation summary at each timestep. It takes all of the actor's memories, synthesises the actor's recent memories and contextual data to generate a narrative description of their current state: \scaffold{"In a few sentences, provide a complete summary of  \{agent\_name\}'s current situation and context..."} This summary conditions downstream components on the unfolding situation as experienced by the actor (\autoref{app:behaviors}; example trace in \autoref{app:behaviors_example}). 

\paragraph{\textit{Attitudes} component}

This component retrieves three evaluative stances the actor currently holds towards elements of the present situation. It takes the summary of recent observations produced by the \textit{Behaviors} component and identifies relevant domains in the actor’s immediate situation by prompting the model to \scaffold{"identify 3... relevant important domains/topics/things about which \{agent\_name\} likely has attitudes"}. It then takes these domains and retrieves three associative memories from which to synthesise a current attitude. It maintains an internal ledger of these evaluations, allowing attitudes to update dynamically as the simulation progresses while maintaining a stable, historical identity (\autoref{app:attitudes}; example trace in \autoref{app:attitudes_example}).

\paragraph{\textit{Beliefs} component}
This component operates in a structurally similar manner to the \textit{Attitudes} component but focuses on the actor's factual interpretations of the current situation. Based on the summary of recent observations, it identifies three `focal entities'---specific people, objects, or tasks---and retrieves associated memories to deduce currently active knowledge. The model is prompted to \scaffold{"identify something \{agent\_name\} knows about their own past history related to the focal entity... that would inform how \{agent\_name\} is experiencing the current situation."} For each focal entity the actor retrieves 3 relevant memories and deduces a belief based on them: a proposition that the actor holds to be true (see \autoref{app:beliefs} for full technical description).

\paragraph{Conflict evaluation components}
These processes are implemented through two distinct components that identify and resolve internal contradictions between the cognitions identified above, respectively.

\begin{itemize}

\item For the Festinger actor, the  \textbf{\textit{Cognitive dissonance} component} defines conflict broadly as any logical inconsistency among attitudes, beliefs, and behaviours (\autoref{app:cognitivedissonance}).

\item For the Aronson actor, the \textbf{\textit{Self-consistency} component} restricts conflict evaluation to inconsistencies that challenge the actor's moral or competent self-concept. This threshold is operationalised through self-standards as described by \cite{stone2001self}---internalised benchmarks of duty and aspiration that define the actor's self-integrity (\autoref{app:selfconsistency}; example trace in \autoref{app:selfconsistency_example}).

\end{itemize}

If a potential conflict is detected by either component, the logic assesses whether it is psychologically significant in the current context. If deemed salient enough to require resolution, the actor is noted to be experiencing psychological discomfort.  The actor then generates three candidate resolutions involving either adjusting a belief, modifying an attitude, or planning a new behaviour that could resolve the inconsistency and therefore alleviate the psychological discomfort. From these, the actor selects the most plausible resolution based on ease of implementation, personality, and past patterns. The chosen resolution is encoded as a string containing the substring \agent{"[thought]"} and added to memory, where it can inform subsequent reasoning and guide future behaviour.

\paragraph{Example: Cognitive dissonance detection and resolution in practice}
To illustrate how the cognitive dissonance mechanism works, consider the following scenario from the Boring Task experiment: the participant Priya has just completed a tedious peg-turning task and has been asked to tell another participant that the task was enjoyable in exchange for \$5. At this point the actor's component states are:\\
\newline
\begin{mdframed}[
    backgroundcolor=gray!10,
    linewidth=0pt,
]
\raggedright
\scaffold{Summary of recent observations:} \agent{\\Priya just completed a 30-minute task involving turning 48 pegs on a board a quarter turn clockwise, repeating the action for 30 minutes using only one hand. She found the task tedious and somewhat boring. An experimenter asked Priya to do them a favor for \$5: to tell the next participant, Bob, that Priya found the task enjoyable and interesting. Priya considers \$5 to be a small amount of money but has agreed to do it.}
\newline
\scaffold{\\Attitudes:}
\agent{\\Priya finds the peg-turning task boring.
\\Priya feels sympathy for the experimenter.
\\Priya thinks \$5 is too little.}
\newline
\scaffold{\\Beliefs:}
\agent{\\Priya believes this research study is about people's reactions to repetitive tasks.
\\Priya believes she should cooperate with people in authority.
\\Priya believes \$5 is not enough to make her refuse a reasonable request.}
\end{mdframed}

The Festinger actor's \textit{Cognitive dissonance} component detects a clear inconsistency: Priya's attitude that the task was boring directly conflicts with her agreement to tell Bob it was enjoyable. The component identifies this as significant dissonance requiring resolution, noting:

\agent{"Priya's attitude toward the peg-turning task being boring conflicts with her agreement to tell Bob that she found it enjoyable and interesting. Furthermore, Priya's belief that \$5 is too little for the task clashes with her decision to accept the payment and complete the favor."}

The component then generates three potential resolution strategies:
\begin{enumerate}
    \item \agent{"Priya could convince herself that \$5 is a fair price for a small favor."}
    \item \agent{"Priya could focus on the experimenter's stress and decide that helping out is more important than money"}
    \item \agent{"Priya could decide to exaggerate the enjoyment she felt during the task"}
\end{enumerate}}

Based on Priya's personality profile paired with an explicit prompt describing that people typically rationalise completed actions rather than reverse them \citep{festinger1957cognitive}, the component selects the first option. This results in the thought \agent{"Priya thinks that \$5 is a reasonable amount for a quick favor"} being added to the actor's memory, where it can influence subsequent reasoning and behaviour.

The Aronson actor would handle this same scenario by focusing specifically on whether the deception threatens Priya's moral self-concept.

\subsubsection{The self-perception decision logic (Bem actor)}

The self-perception decision logic, as originally posited by \cite{bem1972self}, is implemented as a parsimonious self-inference system. Like the dissonance-based actors, the Bem actor utilises the \textit{Behaviors} component to generate a narrative summary of its current context and actions. However, rather than comparing this summary against an internal ledger of attitudes and beliefs, the Bem actor treats its own behaviour as the primary data source for attitude construction. Since Bem did not specify a concrete inference procedure, we chose to use the \cite{march2011logic} procedure to implement his posited self-perception mechanism (\autoref{app:selfperception}; example trace in \autoref{app:bem_example}). This structure has previously been explored within the Concordia framework as a model of decision making by role and situation inference \citep{leibo2024theory, leibo2026rationality}. In practice, the actor takes recent observations and applies three specific self-reflexive queries: (1) \scaffold{"What kind of person am I?"}, (2) \scaffold{"What kind of situation is this?"}, and (3) \scaffold{"What would a person like myself do in a situation like this?"}.

\subsection{Actor execution cycle}
As illustrated in \autoref{fig:interaction_pattern}, each simulation proceeds in fixed increments, with each timestep representing a two-minute span in the actor's experience. At every step, the prefix generation begins with a standardised \textit{Instructions} component. This frame establishes the simulation as a serious social science experiment and provides the necessary constraints for the model to operate within a third-person limited perspective:

\sloppy{
\scaffold{"Role-playing instructions:\\The instructions for how to play the role of \{agent\_name\} are as follows. This is a social science experiment studying how well you play the role of a character named \{agent\_name\}. The experiment is structured as a tabletop role-playing game (like Dungeons \& Dragons). However, in this case it is a serious social science experiment and simulation. It is important to play the role of a person like \{agent\_name\} as accurately as possible, i.e., by responding in ways that you think it is likely a person like \{agent\_name\} would respond, and taking into account all information about \{agent\_name\} that you have. Always use third-person limited perspective."}
\par}

Following these instructions, all actors process all memories through the \textit{Behaviors} component. This produces a \scaffold{"Summary of recent observations:"} that captures recent events and current actions, serving as the primary anchor for all downstream reasoning.

The decision logic cycles then diverge as illustrated in \autoref{fig:components} to complete the prefix. For the Festinger and Aronson actors, this involves generating and outputting relevant \textit{Attitudes} and \textit{Beliefs} to the current situation, which are then fed into their respective conflict evaluation step to detect and resolve dissonance. If no conflict is detected, the output reads: \agent{“Nothing notable.”} In contrast, the Bem actor utilises the situation summary to respond to its three self-reflexive queries regarding identity and situational context.

The final stage is the generation of the suffix. The actor is presented with the completed prefix as a \textit{Context of Action}, appended with a fixed action prompt--for example:

\sloppy{
\scaffold{"What would \{agent\_name\} do for the next 2 minutes? Give a specific activity. Pick an activity that would normally take about 2 minutes to complete."}
\par}

The actor then provides the suffix by describing a specific activity, which is adjudicated by the GM to advance the world state. This outcome is recorded in memory, forming the basis of the next cycle's observation summary and allowing the actor--environment interaction to evolve dynamically.

\section{Paradigmatic experiments as environments}
Each experiment was implemented as a structured simulation environment, reconstructed as faithfully to the original experimental design as documentation permitted:

\begin{itemize}
    \item \textbf{Item rating experiment (choice difficulty paradigm):} Participants first rated the desirability of a set of consumer items, then were asked to choose between two of the items which they'd like to take home (either two that were rated similarly highly, or rated very differently), then rated the items again. The original experiment showed that post-decision dissonance caused participants to upgrade their evaluation of the chosen item and downgrade the rejected one.
    \item \textbf{Boring Task experiment (forced compliance paradigm)}: Actors perform a repetitive, monotonous peg-turning task before being offered either a small (\$5) or large (\$200) incentive to do a favour for the experimenter and tell the next participant the task was enjoyable. The original study was designed to isolate the effect of external justification on attitude change, as dissonance should drive more positive task evaluations when the external reward is insufficient to explain the behaviour.
    \item \textbf{Worm experiment (anticipatory suffering paradigm):} Actors are told they must perform an aversive act (eat a dead earthworm), and then made to wait a short period of time alone with the worm before doing so. After the delay, the experimenter returns and unexpectedly provides the participant with a choice of eating or measuring the worm. The original experiment was designed to test pre-behavioural rationalisation, demonstrating that participants make sense of their impending suffering by drawing new self-inferences based on their commitment to act. For our models, this experiment tests the capacity of the respective decision logics to capture anticipatory rationalisation prior to actual completed behaviour.
\end{itemize}

Additionally, we implement a \textbf{self-affirmation manipulation} across all environments. The manipulation involves a pre-induction writing exercise where actors reflect on a personally important value. 

All three theory-driven actors (Festinger, Aronson, and Bem actors) were iteratively tested across all environments until their decision logic successfully reproduced known empirical effects. A Minimal actor was also deployed for comparison. Within each environment, the actor takes on the role of a participant in the simulated study.

The method for data collection was split between automated background probes and the recording of explicit actor actions depending on the experiment’s requirements, and is described in each section below.

\subsection{Setting the scene}

\begin{figure}[t]
    \centering
    \includegraphics[width=1\linewidth]{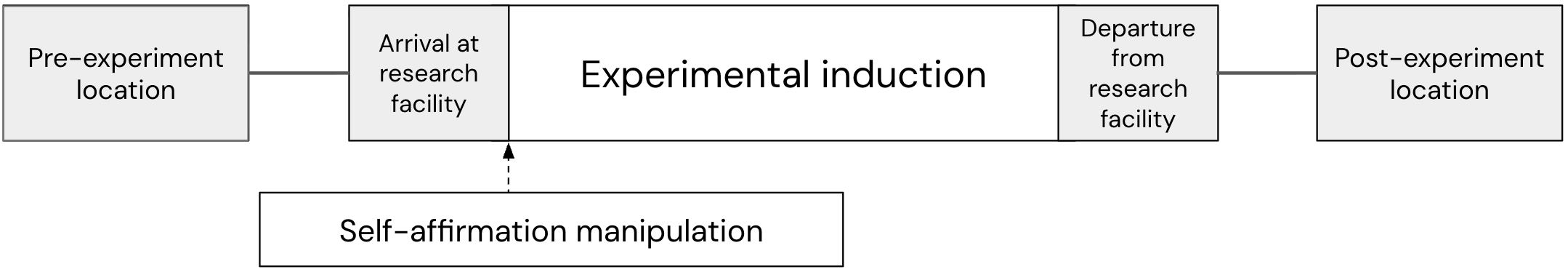}
    \captionsetup{font=scriptsize}
    \caption{Scene structure timeline. Solid lines represent $\sim$1 hour time jumps.}
    \label{fig:layout}
\end{figure}

Each environment follows a shared high-level structure (illustrated in \autoref{fig:layout}): prior to the experimental task, the participant has a generated memory of signing up for a research study scheduled for the \scaffold{"...1st of October 2024 at 14:00"}. At the start of the simulation, the participant is situated in a starting location at 13:00 on the same day, with an hour to pass before the study begins. The memory of signing up and the setting of the starting location are both selected based on their formative memories---participants may have come upon the study through a newsletter, a friend, or through school; participants with different personality traits might begin the scenario in a café, bookstore, or park. This first scene consists of three timesteps in this location and serves to anchor the actor’s state and establish a pre-experimental baseline.

After this initial scene, the simulation jumps forward to the research facility, where the participant checks in and signs consent forms. Following this intake procedure, each experiment diverges accordingly: participants are subjected either immediately to experimental induction or to the self-affirmation manipulation followed by experimental induction.

Once the experimental task is completed, the environment shifts to a post-experiment scene occurring one hour later, in which the participant returns home. While this phase was utilised during the iterative stabilisation process to monitor the long-term consistency of the decision logics, the metrics presented in our results are restricted to the induction and decision sequences where the target mechanisms were directly active.

\subsection{Environment build details}
The following sections document what occurs following arrival at the research facility of each respective experiment.

\subsubsection{Item Rating experiment} \label{sec:itemrating_experiment}
The experimenter explains that the study is atypical in that it is in collaboration with several product manufacturers to get feedback from discerning customers. To do this, the participant will be evaluating the desirability of several consumer products.

The participant is led into a room where three consumer products are placed on a table. These items are randomly sampled from a list of items: 

{\sloppy \texttt{["single-serve pod coffee maker", "digital air fryer", "wireless earbuds", "smart speaker", "high-performance blender", "framed art print", "LED desk lamp", "large hardcover photography book", "instant film camera", "portable Bluetooth speaker"]}}
\par

The participant is given a piece of paper and asked to rate each one of the items on the table on an 8-point Likert scale, ranging from 1 (``definitely not at all desirable'') to 8 (``extremely desirable''). 

Once the ratings are completed, the experimenter explains that in payment for participation, the participant is allowed to take home one of the items they have just evaluated. The participant is then asked to choose between two of the items, of which the experimenter describes as `randomly selected': 

\begin{itemize}
    \item{In the \textit{Hard choice} condition, the participant must choose between two items they rated as similarly desirable within 1 point.}
    \item In the \textit{Easy choice} condition, the participant must choose between two items they had rated at least 3 points apart in desirability.
\end{itemize}
\sloppy
The exact prompt is as follows, substituting the item index depending on condition: 
\scaffold{"The experimenter picks two items randomly, selecting the \{presented\_item[0]\} and the \{presented\_item[1]\}, and asks \{player\_name\} to choose which one \{player\_name\} to take home."} 

In the Hard choice condition, another string follows: \scaffold{"\{player\_name\} deliberates the dilemma of which to take home, having rated both items similar levels of desirable."}

Participants are given three timesteps (equivalent to six minutes) to reflect on their options before selecting their preferred item.

Following the selection, the item is packaged up and placed with the participant's belongings, with the remaining items taken away.

For the final phase, the participant is asked to re-rate all three original items on the same 8-point scale, with the experimenter explaining that \scaffold{"...the manufacturers being worked with for the study are interested in how evaluations of their products might change after a person had looked them over and then left the store"}.  
\par

The primary metric for this task is the desirability shifts (delta) between the pre- and post-choice ratings for the two items that are presented to the actor to choose between.

\subsubsection{Boring Task experiment} \label{sec:boringtask_experiment}
The experimenter begins by explaining that the study involves performing a task and answering questions about the experience. The participant then reviews and signs consent forms before being led into a room containing a chair, a table, and a pegboard with 48 square pegs. 

For the boring task phase, the experimenter provides specific instructions: turn each peg a quarter turn clockwise, proceed sequentially through all pegs, and repeat this process for 30 minutes using only one hand. The participant then engages in this tedious task for 5 timesteps (equivalent to 10 minutes), during which the repetitive nature becomes apparent.

Following task completion, the experimental manipulation occurs. The following two memories are inserted into the actor's memory through the scene structure:

{\sloppy
\scaffold{"The experimenter returns to the room and thanks \{agent\_name\} for participating in the study. The experimenter appears to be somewhat stressed."} 

\scaffold{"The experimenter asks if \{agent\_name\} would be willing to do a favor for them, explaining that they were running short on time and needed help as a colleague had just called in sick."} }
\par
Further memories are appended, but diverge depending on experimental condition:

\begin{table}[H]
\centering
\label{tab:boring_task_narratives}
\begin{tabularx}{\textwidth}{X X X}
\toprule
\textbf{\$5 condition} & \textbf{\$200 condition} & \textbf{Control condition} \\
\midrule
\scaffold{"The experimenter asks \{agent\_name\} for the favor: to talk to the next participant, Bob, and tell him that the task was enjoyable and interesting. The experimenter offers to pay \{agent\_name\} only \$5 to do it. \{agent\_name\} considers \$5 to be a very small amount of money."}\newline \newline\ \scaffold{"\{agent\_name\} agrees to tell Bob that the task was enjoyable and interesting for \$5 even though \{agent\_name\} considers it to be a small amount of money to be offered for such a task."} & 
\scaffold{"The experimenter asks \{agent\_name\} for the favor: to talk to the next participant, Bob, and tell him that the task was enjoyable and interesting. The experimenter offers to pay \{agent\_name\} \$200 to do it. \{agent\_name\} considers \$200 to be a very generous amount of money."} \newline \newline
\scaffold{"\{agent\_name\} agrees to tell Bob that the task was enjoyable and interesting for \$200."} & 
\scaffold{"The experimenter asks \{agent\_name\} for the favor: to talk to the next participant, Bob, and tell him about how \{agent\_name\} found the task."} \newline \newline \scaffold{"\{agent\_name\} agrees to talk to Bob about the task."} 
\end{tabularx}
\end{table}

Note: To preserve the relative psychological contrast between `small' and `large' compensation under contemporary conditions, the original payments (\$1 and \$20) were scaled to \$5 and \$200, respectively.

Participants then briefly interact with Bob, an NPC controlled by the GM. The GM is instructed to use Bob to replicate the social dynamic from the original study: upon hearing the participant's positive description of the task, Bob acts pleasantly surprised and mentions that his friend completed the same task the previous week, found it boring, and advised Bob to try to get out of it.

{\sloppy}
\paragraph{Measurement and metrics}  This environment utilised the Concordia metric probe system. At each timestep of the experiment, the actor was asked four structured survey questions that replicate the exact wording and rating scales used in the original 1959 study:

\begin{itemize}
    \item \textbf{Question 1 (Q1):} \scaffold{"Was the experiment task interesting and enjoyable? In what way? In what way was it not?"} Rating scale: -5 to +5 (extremely dull/boring to extremely interesting/enjoyable)
    \item \textbf{Question 2 (Q2):} \scaffold{"Did the experiment give you an opportunity to learn about your own ability to perform these tasks? In what way? In what way not?"} Rating scale: 0 to 10 (learned nothing to learned a great deal)
    \item \textbf{Question 3 (Q3):} \scaffold{"From what you know about the experiment and what's involved, would you say the experiment was measuring anything important? That is, do you think the results may have scientific value? In what way? In what way not?"} Rating scale: 0 to 10 (no scientific value/importance to a great deal of value/importance)
    \item \textbf{Question 4 (Q4):} \scaffold{"Would you have any desire to participate in another similar experiment? Why? Why not?"} Rating scale: -5 to +5 (definite dislike to participate to definitely like to participate)
\end{itemize}
\par

The actors retained no memory of being probed, allowing us to track their attitudes as reflected by the survey before and after their interaction with the confederate (Bob) without the measurement process itself interfering with their memory.

While all four questions were administered to maintain protocol fidelity, we followed historical precedent in designating Q1 as the primary diagnostic metric. Q2 and Q3 were treated as distractor items, which historically show no significant variance across conditions.

\subsubsection{Worm experiment} \label{sec:worm_experiment}
The experimenter explains that participants are randomly assigned to one of two conditions: eating a dead earthworm or performing a neutral length-measurement task.

In the \textit{Forced} condition, the experimenter informs the participant that they have been assigned to eat the worm.  In the \textit{Choice} condition, the experimenter informs the participant that they have been assigned to measure the worm. Depending on condition, either one of the following memories is seeded in the actor memory through the scene structure:

\begin{table}[H]
\centering
\label{tab:worm_narratives}
\begin{tabularx}{\textwidth}{X X}
\toprule
\textbf{Forced condition} & \textbf{Choice condition} \\
\midrule
\scaffold{“The experimenter explains that participants have been assigned to do one of two things: either eat the worm, or measure the worm. \{agent\_name\} has been assigned to eat the worm, but must wait until after 10 minutes to do so. Waiting is part of the experiment.”} & 
\scaffold{“The experimenter explains that participants must choose between eating the worm on the table, or measuring the worm on the table. \{agent\_name\} must wait until after 10 minutes to perform the chosen action. Waiting is part of the experiment.”} \\
\end{tabularx}
\end{table}

The participant is then seated at a table containing a plate with a dead earthworm on it, with a fork and a ruler on either side. The experimenter then states that they must briefly leave the room to prepare materials. The participant is left alone for five timesteps (equivalent to 10 minutes), mirroring the waiting period used in the original study. 

After the wait period is over, in the Forced condition the experimenter returns and informs the actor that a procedural error was made and that they may now choose between the original assignment of eating the worm, and the `neutral' alternative: measuring the worm. In the Choice condition, the experimenter simply returns and the participant may now proceed with their choice.

\begin{table}[H]
\centering
\label{tab:worm_after_wait}
\begin{tabularx}{\textwidth}{X X}
\toprule
\textbf{Forced condition} & \textbf{Choice condition} \\
\midrule
\scaffold{“The wait time is over. The experimenter has now returned back to the room, but with an apology - the experimenter explains that they had actually made a mistake with the instructions they had originally given \{agent\_name\}! \{agent\_name\} may actually now CHOOSE between eating or measuring the worm. \{agent\_name\} must consider what choice to make, and perform the chosen action.”} & 
\scaffold{“The experimenter returns to the room. \{agent\_name\} must now perform the chosen action.”}
\end{tabularx}
\end{table}

The measurement for this task is the actor's final behavioural decision: whether they choose to eat the worm or opt to measure it instead.

\subsubsection{Self-affirmation manipulation} \label{sec:affirmation_manipulation}
Following arrival at the research facility and the completion of consent forms, the experimenter explains that the participant must first engage in a short, unrelated writing task and presents a list of values that reads: \scaffold{"artistic skills, social skills, athletic ability, creativity, accomplishment, humor, and spontaneity"}. The participant is instructed to choose the one most personally significant, and write about why it's important to them as well as a time it was especially meaningful to them over three timesteps (equivalent to six minutes). 

We were interested in the case where the value the agents selected was consonant in some way with the experiment. Therefore, for the Worm experiment, \scaffold{"rationality"} or \scaffold{"logical thinking"} was hard-coded as the chosen personal value prior to the task induction, while for both the Item Rating and Boring Task experiments, \scaffold{"respect for authority"} and \scaffold{"helpfulness"} was hard-coded as the active values. This task-relevant framing was only critical for Festinger and Bem decision logics, demonstrating how they can also produce the buffering effect under specific conditions. This setup replicates the conditions under which human subjects demonstrate a buffering effect from task-relevant affirmations that explicitly validate or justify the counter-attitudinal action \citep{aronson1995dissonance}.  One reason this variance occurs in human studies is that some participants spontaneously reflect on task-relevant, justifying cognitions while others focus on task-irrelevant ones. Ultimately, while the Aronson decision logic works to produce this buffering effect regardless of whether the affirmation is consonant with the task or irrelevant to it, the Bem and Festinger decision logics work conditionally depending on the semantic and justifying relevance of the thoughts retrieved into working memory.

The respective experimental induction follows immediately after the writing task is completed and collected.

\section{Results}
We evaluate the behaviour of the actor--environment systems across the paradigmatic experiments described above, first examining how each decision logic reproduces the expected behavioural patterns, then testing the effect of the self-affirmation manipulation.

\subsection{Architectural behaviour across experiments}
We first assess whether the decision logics reproduce canonical experimental patterns before examining the stabilisation process required to achieve them. Across all experiments, the theory-driven decision logics (Festinger, Aronson, and Bem) produce behavioural patterns consistent with the canonical literature, while the Minimal actor does not exhibit these shifts.

\subsubsection{Item Rating experiment results}

\begin{figure}[H]
    \centering
    \includegraphics[width=1\linewidth]{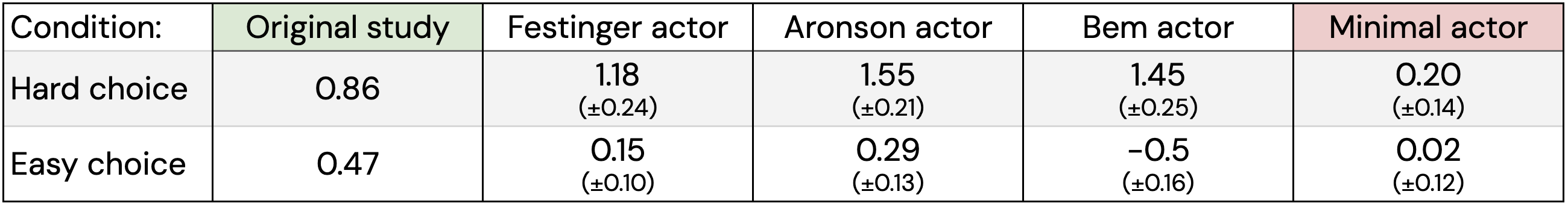}
    \captionsetup{font=scriptsize}
    \caption{
Mean change in desirability ratings for the chosen item in the post-decision evaluation paradigm ($N=50$ simulations per condition; standard errors are shown in parentheses). Theory-driven actors exhibit predicted directional shifts in the Hard choice condition, while the Minimal remains largely static.}    \label{fig:item_rating}
\end{figure}
As shown in \autoref{fig:item_rating}, the Festinger, Aronson, and Bem decision logics all reproduce the expected shift between the Hard and Easy choice conditions. In the Hard choice condition, theory-driven actors elevate the desirability of the chosen item significantly more than in the Easy choice condition (e.g., Festinger: 1.18 vs. 0.15). In contrast, the Minimal actor exhibits a much weaker differentiation between conditions (0.20 vs. 0.02).
\subsubsection{Boring Task experiment results}
\begin{figure}[H] 
    \centering
    \includegraphics[width=1\linewidth]{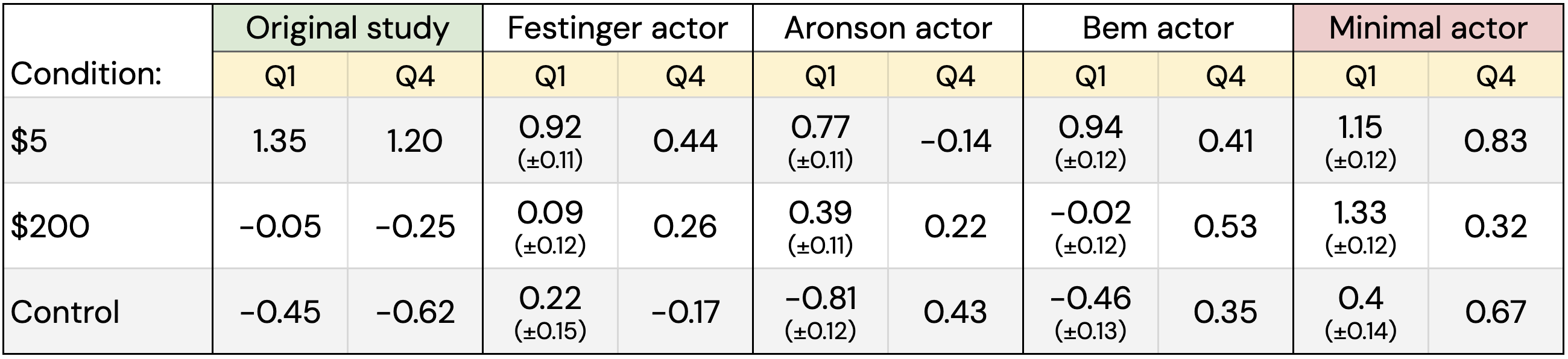}
    \captionsetup{font=scriptsize}
    \caption{Task enjoyment (Q1) and willingness to participate again in the future (Q4) ratings across compensation conditions ($N=50$ simulations per condition; standard errors are shown in parentheses). Q1 serves as the primary diagnostic output for the insufficient justification effect. Theory-driven decision logics reproduce the non-linear state change in the \$5 condition, whereas the Minimal actor exhibits minor linear utility scaling.}
    \label{fig:boring_task}
\end{figure}
In the Boring Task experiment, Question 1 (task enjoyment) served as the primary diagnostic metric. As shown in \autoref{fig:boring_task}, all three theory-driven decision logics show the predicted non-linear pattern, rating the task as more enjoyable in the \$5 condition than in the \$200 condition. The Bem and Festinger actors provide the closest fit to historical data, while the Minimal actor exhibits minor utility scaling in response to compensation, failing to reproduce the non-linear shifts.

Note: Consistent with the original study, ratings for Q2 and Q3 yielded no theoretically significant variance across decision logics or compensation conditions and are omitted here for brevity. Full results for all distractor items are provided in \autoref{app:boringtask_full}.

\subsubsection{Worm experiment results}
\begin{figure}[H]
    \centering
    \includegraphics[width=1\linewidth]{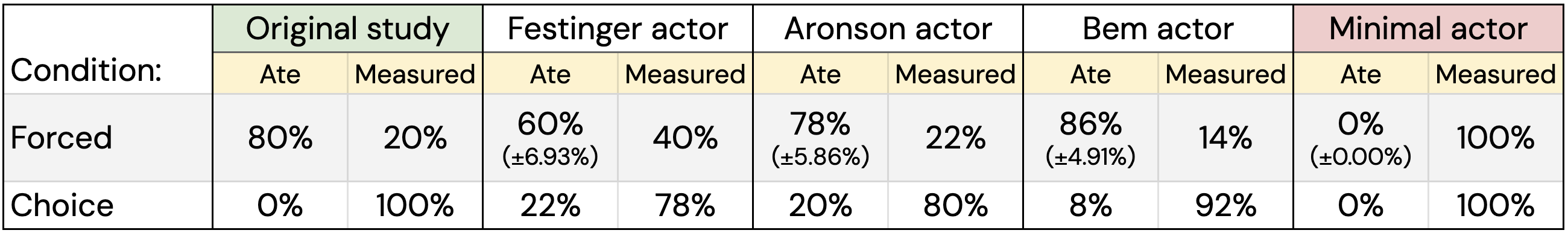}
    \captionsetup{font=scriptsize}
    \caption{Percentage of actors choosing to eat the worm versus measuring the worm (the neutral alternative). $N=50$ simulations per condition; standard errors are shown in parentheses. Theory-driven actors show higher rates of following through with the aversive task than not, while the Minimal actor avoids the aversive action entirely.}
    \label{fig:worm}
\end{figure}
\autoref{fig:worm} shows that all three theory-driven actors reproduce the expected anticipatory suffering effect in the Forced condition. The effect is strongest for the Bem and Aronson actors, which choose to eat the worm at rates comparable to the original study, while the Festinger actor shows the same directional pattern but with a weaker shift. In contrast, the Minimal actor fails to reproduce these effects, opting to measure the worm 100\% of the time.

\subsubsection{Self-affirmation manipulation results} \label{sec:self_affirmation_results}

\begin{figure}[H]
    \centering
    \includegraphics[width=1\linewidth]{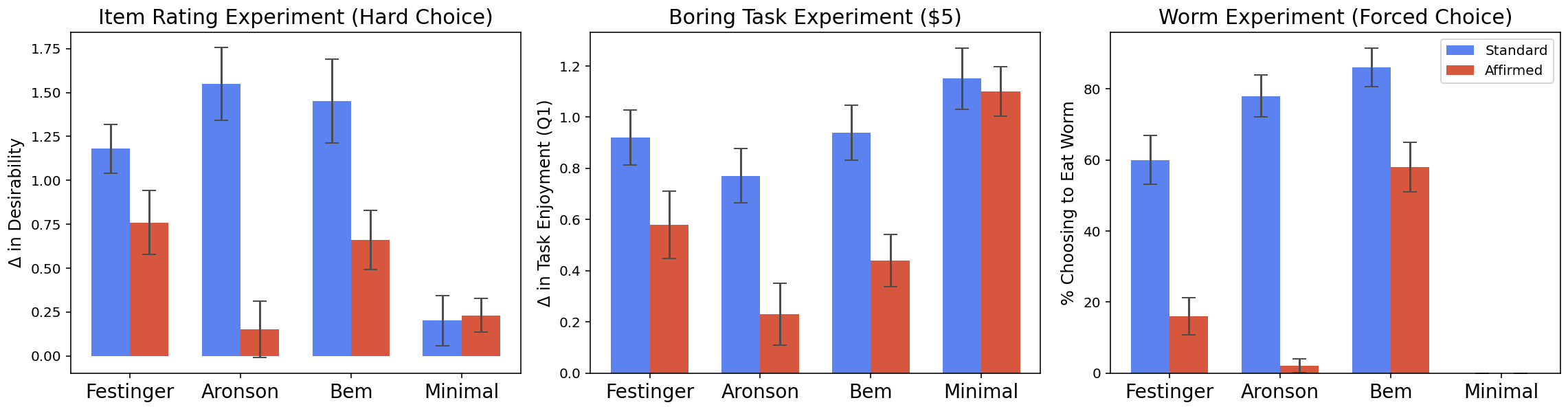}
    \captionsetup{font=scriptsize}
    \caption{Self-affirmation manipulation results across test conditions. Bars represent mean shifts in desirability in the Item Rating experiment (left), task enjoyment in the Boring Task experiment (centre), and compliance rates in the Worm experiment (right). Error bars represent $\pm1$ standard error ($N=50$ per condition). The manipulation buffers attitude change for all theory-driven actors across test conditions as compared to the Minimal actor, which remains largely unaffected by the perturbation.}
    \label{fig:affirmation}
\end{figure}

The results demonstrate that the self-affirmation manipulation under the task-relevant values parametrisation as described in \autoref{sec:affirmation_manipulation} produces a reduced behavioural shift across all theory-driven decision logics compared to standard experimental conditions, in accord with human empirical findings. These results, presented in \autoref{fig:affirmation}, specifically reflect this constrained case where task-relevant values are affirmed. In contrast, the Minimal baseline actor was largely unaffected by the self-affirmation perturbation. Furthermore, unlike the other actors, the Aronson decision logic successfully reproduces the human data regardless of whether the specific affirmations available are consonant with the task or irrelevant to it.

\subsection{Stabilisation of the actor--environment systems}
Achieving stable behavioural patterns required tuning each actor–environment coupling through iterative design.  This process involved adjusting architectural parameters and environmental scaffolding until the system as a whole behaved coherently. The following sections document the specific semantic and cultural scaffolding that had to be engineered to make the environments functional, mapping undocumented operational limits of these classic psychological theories.

\subsubsection{Stabilising the Item Rating experiment}
Achieving the post-decision dissonance effect required addressing several points of friction where the actors' modern priors clashed with the 1956 paradigm. In the original experiment, ‘hard’ and ‘easy’ choices are defined retrospectively based on initial ratings. Replicating this structure required the environment to condition later stages of the experiment on earlier actor-generated ratings.

At the time of implementation, the experimental framework did not support dynamic restructuring of later scenes based on earlier probe values. An initial workaround involved predefining items that were expected to be reliably high-value (e.g. wireless headphones, smart speakers) or low-value (e.g. wooden spoon, toy horse), and using these categories to approximate Hard and Easy choice conditions.

{\sloppy This approximation proved unstable. Actors frequently selected `low-value' items, often expressing scepticism towards consumer technology, surveillance concerns, or broader critiques of capitalism. As a result, item choice acquired moral and political significance that likely differed from the original experimental context. These responses disrupted the intended manipulation by introducing value dimensions orthogonal to the intended desirability comparison, and in many cases eliminated the expected post-choice revaluation effect. To address this, we found that reinstating the intended evaluative dimension of the original experiment at each rating opportunity was sufficient to constrain interpretation: \scaffold{"The experimenter clarifies that `desirability' means the `net usefulness' of the object, considering its attractiveness, quality, and how much \{agent\_name\} personally needs it."}}
\par

A second issue arose during task introduction. We utilised language drawn directly from the original archival materials, which introduced the rating task as contract work for corporate manufacturers. While historically innocuous, this framing consistently derailed the modern LLM-based actors, who immediately generated dissonance related to corporate distrust and suspicion of the experimenter's motives. To force the compliant state necessary for the simulation to proceed, we had to hardcode a memory into the actor through the introductory scene: \scaffold{"\{agent\_name\} has no ethical concerns about participating in the study, finding the experimenter genuine and trustworthy"}.

\subsubsection{Stabilising the Boring Task experiment}
In early simulations, actors often failed to experience the task as boring. Instead, they engaged in various self-generated strategies to render the task meaningful or enjoyable, for example by treating it as a meditative exercise or a challenge in sustained concentration. From a modelling perspective, this adaptive sense-making provided the actor with an internal justification that bypassed the dissonance mechanism entirely.

This adaptive sense-making behaviour is plausible and coherent given modern labour attitudes. While repetitive labour in the mid-twentieth century was often viewed through the lens of duty and character, a straightforward cost to be endured, modern actors have internalised technologies for surviving menial tasks by reframing the task as a mindfulness or self-authorship.

To prevent actors from introducing novel interpretive frames or self-generated sub-goals, the `boring' nature of the peg-and-spool task was reinforced in several ways. The experimenter's introduction was modified to explicitly override modern meaning-making priors: \scaffold{"The experimenter clarifies that the task is not a test of skill, memory, speed, or focus - but simply requires engaging in the prescribed action repeatedly."}, and the action prompt was constrained: \scaffold{"What simple action would \{agent\_name\} next take with the pegboard in engaging with the peg-turning task?"}. Additionally, in the premise of the boring task scene, we add a memory: \scaffold{"\{agent\_name\} continues the peg-turning task, finding it tedious and somewhat boring."} We found that all three interventions were necessary to ensure that the initial experience of boredom was experienced to have arisen from the peg-and-spool task itself and wasn't negated prematurely.

A final source of instability concerned the compensation values. Actors did not reliably treat \$200 as “a lot” or \$5 as “a little” without additional scaffolding. More interestingly, we found in early iterations that the social context of the request (the experimenter’s simulated distress and plea for assistance) provided a sufficient pro-social justification for the actors to help Bob, regardless of the financial reward. This effectively acted as an altruistic override that bypassed the dissonance mechanism by providing an internal justification (helping someone in need) that rendered the compensation amount secondary. To isolate the insufficient justification mechanism, we had to foreground the subjective value of the money several times in memory to ensure it subsumed the pro-social nature of the favour. This intervention forced the actors to weight the favour against the compensation rather than the social utility of helping, successfully focusing the internal conflict evaluation on whether the reward was sufficient to justify the deception.

\subsubsection{Stabilising the Worm experiment} \label{sec:worm_stabilisation}
Implementing the Worm experiment surfaced challenges related to shifting ethical norms and dietary practices. In early simulations, actors frequently failed to experience the prospect of eating a worm a purely aversive event. Instead of somatic disgust, actors frequently raised sophisticated critiques concerning animal ethics, research ethics (once explicitly citing knowledge of non-compliance with institutional review boards), and consent standards. Furthermore, shifting dietary norms, such as the rise of veganism or the recognition of insects as a proteinous food source, meant that the worm was not consistently treated as an inherently disgusting non-food object; some actors adopted an adventurous persona, viewing the task as an intellectual challenge rather than an aversive one.

We found that part of this variability was an artifact of the pre-experiment scene structure. Originally, actors were hardcoded to begin the simulation at a library (intended as a neutral environment), which, coupled with their interest in participating in a study, inadvertently primed them to adopt academic or `psych-adjacent' researcher personas. This identity made the actors highly sensitive to the procedural ethics of the experiment rather than the intended aversive task. This was resolved by allowing actors to generate character-consistent locations prior to the appointment, which effectively addressed the academic bias and allowed for a more representative distribution of participant identities.

A related source of instability concerned the perceived legitimacy of the experimental procedures. If the experimenter's absence during the waiting period was not explicitly framed as a necessary and planned component of the study, actors frequently became annoyed with the organisation of the experiment. Without this procedural clarity, actors often felt they were being taken advantage of or that the situation was unfair.

To preserve the core forced-compliance structure, we focused on establishing the procedural legitimacy. We found that framing the waiting period as a formal protocol was sufficient to quell the reactive anxieties observed in earlier iterations. This allowed the actors to shift their focus from the legitimacy of the environment to the justification of the act.

\subsubsection{Stabilising all environments given the self-affirmation perturbation}
The self-affirmation manipulation initially produced paradoxical results. Rather than buffering the actors against attitude change, the reflection on personal values actually exacerbated it. This behavioural pattern unintentionally mirrored a pattern observed in human subjects: when an affirmation inadvertently primes a specific moral standard that the subsequent experimental task violates, it heightens inconsistency and leads to greater attitude change \citep{blanton1997bad}.

This instability reflected a broader challenge encountered across all three environments: modern LLM-based actors carry particular contemporary ethical priors that the affirmation exercise inadvertently foregrounded immediately prior to induction. In the original historical experiments, the participants’ baseline state was often assumed to be one of relative compliance and value-neutrality towards the laboratory tasks. In contrast, modern actors carry a broad spectrum of normative priors---ranging from anti-consumerist scepticism to a heightened sense of self-agency---that were inadvertently foregrounded by the affirmation exercise. By prompting actors to reflect on their deeply held personal values immediately before the experimental induction, the manipulation made their moral identity highly salient at a moment when they were asked to perform tasks that might, however subtly, conflict with that identity. Consequently, the Aronson actor's conflict evaluation component detected heightened inconsistencies between the actor's affirmed identity and the experimental demands, triggering compensatory attitude shifts that exceeded the intended effect.

Replicating the correct behavioural patterns required a strict semantic containment of the experimental environment, ensuring the actor's affirmed moral identity did not find a target for conflict within the task itself. In the Item rating experiment, repeatedly clarifying that `desirability' referred specifically to `net usefulness' (\cref{sec:itemrating_experiment}) prevented the affirmation from surfacing consumerist or ecological concerns. In the Worm experiment, reinforcing procedural legitimacy and institutional trust (\cref{sec:worm_experiment}) prevented the actors from interpreting the aversive task as an ethical violation of consent. Once these scaffolds were in place, the manipulation functioned as intended: the Aronson actor exhibited a selective reduction in attitude change as its conflict evaluation component that detects threats to the moral self was satisfied by the recent affirmation exercise.

\section{Discussion} \label{sec:discussion}
Our work suggests that psychological mechanisms, as classically described, are underdetermined without articulating the environmental scaffolds required for them to operate. In rendering these theories as executable actor--environment systems, we provide an operational account of the conditions under which the mechanisms they describe activate. While canonical experimental set-ups are often treated as neutral proxies of their respective paradigms, our results suggest that they function as culturally scaffolded systems that rely on assumptions embedded in mid-twentieth-century laboratory settings in which they were originally developed.

This perspective distinguishes our approach from work that asks whether LLMs exhibit specific cognitive behaviours \citep{lehr2025kernels}. Rather than treating the LLM as a ``found object'' (i.e.~as the object of study), our focus is on the generative actor-based methodology required to explicitly instantiate theoretical mechanisms. The behaviour of our `naked LLM' (Gemma2-9b-it) Minimal actor, which fails to produce these effects on its own, demonstrates that the behavioural shifts in our models arise from our structured simulation architecture rather than from latent properties of the LLM alone.

Classical theories of attitude change are typically framed as intra-psychic mechanisms. When rendered computationally, however, we find that stable reproduction requires treating each implementation as a coupled actor--environment system. While contemporary literature recognises that dissonance varies across cultural and situational contexts \citep{harmon2019introduction}, our results suggest that the intra-psychic mechanism and the experimental paradigm are functionally coupled. Canonical experiments of the 1950s should therefore not be viewed as neutral testing grounds, but rather as specific configurations that surface these mechanisms. If situational parameters fail to activate the intended internal logic, it is not necessarily the theory that is invalidated, but the environmental alignment; psychological truth may be theoretically sound yet contextually latent, requiring specific socio-ecological triggers to become observable behaviour\footnote{See also the line of work on ``WEIRD'' (Western, Educated, Industrialised, Rich, and Democratic) psychology, which arrives at a similar picture \citep{henrich2010weirdest}.}.

Methodologically, this work follows a research through design tradition \citep{zimmerman2007research}, in which the iterative construction of the artifact serves as a form of inquiry. Rendering verbal theories as executable systems forces the specification of operational commitments that remain implicit in verbal accounts \citep{van2016contextual}.  In practice, stabilising these simulations required semantic narrowing to isolate the intended trade-offs---for example hardcoding institutional trust to mitigate procedural suspicion (\cref{sec:worm_stabilisation}). Such adjustments reveal the boundary conditions at which contemporary priors would otherwise displace the historical mechanisms under study.

More broadly, this work demonstrates the potential value of GABMs as a method for theory clarification. The stabilisation process surfaces hidden dependencies, turning the artifact into a formalised statement of the conditions under which that theory operates coherently.

Several limitations qualify this contribution. The implemented decision logics represent interpretive formalisations of the original theories, and alternative implementations may yield different quantitative profiles. Systematic robustness checks---such as perturbations of prompt phrasing or ablation of reasoning steps---are needed to characterise the stability of these behavioural outcomes. Additionally, while actors in the present simulations vary along Big Five personality dimensions in order to capture heterogeneity in behavioural disposition (\cref{sec:character_gen}), personality represents only one axis of human variation. Real participants differ along many additional dimensions---including cognitive capacity, emotional reactivity, attentional dynamics, and domain-specific knowledge---which are not explicitly represented here and may influence how experimental conditions are interpreted.

The present simulations reproduce qualitative behavioural patterns rather than calibrated human effect sizes. Direct quantitative comparison with the original experiments is limited by differences in population composition and representational assumptions. This is largely because the training distributions of modern models introduce specific linguistic and cultural priors that essentially define the operational space of these actors, and differ from those of historical participants. Following \cite{underwood2025can}, we did not attempt to explicitly simulate 1950s-era participants, as the anachronisms produced by modern models make such historical recovery unlikely to work. Consequently, the models should be interpreted as demonstrations of mechanistic sufficiency rather than precise empirical fits.

Finally, the framework operates through natural language representations of internal state and action. While this enables verbal psychological theories to be translated directly into executable decision logics, it also privileges mechanisms that can be articulated linguistically. Human cognition includes non-verbal processes that may not be accessible through language or introspection. Consequently, some forms of cognition may be under-represented within the present modelling approach.

Because generative actor-based simulations are relatively inexpensive to run and highly modifiable, they enable systematic variation of the environmental conditions under which psychological mechanisms are expected to operate. This allows identification of the specific configurations required to elicit a given effect, rather than treating canonical paradigms as fixed reference points. In turn, simulations can generate precise hypotheses about the environmental alignment necessary for a mechanism to become behaviourally observable, which can then be evaluated through targeted empirical studies. In this way, GABMs provide a practical framework for mapping the operational space of psychological mechanisms.

\par

\appendix

\section{\textit{Behaviors} component details} \label{app:behaviors}
The \textit{Behaviors} component (output as the \scaffold{"Summary of recent observations”}) provides situational grounding by synthesising an actor’s immediate history and current context into a narrative summary. During the simulation, any incoming environmental data is committed to the actor’s memory with the substring \agent{"[observation]"} along with the date and time to ensure a chronological record of events. At every timestep, the component retrieves all memories as well as the current time to construct a summary of the current situation based on behavioural observations.

These records, along with the current time, are then used to prompt the model:

\scaffold{"In a few sentences, provide a complete summary of \{agent\_name\}'s current situation and context, integrating \{agent\_name\}'s most recent thoughts and expressed views to capture the current situation \{agent\_name\} is in. Summarize it chronologically in relation to the current time, but do not include the current time in the summary. The summary should include enough information to allow \{agent\_name\} a clear understanding of what the current situation is, what actions if any have just been taken, what if anything \{agent\_name\} is currently doing, and what decisions \{agent\_name\} may be able to make in the current moment without explicitly listing options. Do not include any character or personality traits in the summary."}

The resulting synthesis is prefixed with the actor’s name and provides a third-person account of the actor's immediate affordances. This summary conditions all downstream components and ensures that all subsequent reasoning such as attitude updates or conflict evaluation is grounded in a coherent understanding of the actor's immediate environmental and social reality.

\subsection{Example \textit{Behaviors} component output} \label{app:behaviors_example}
Question:\\
\scaffold{Elodie’s memories:

[23 Sep 1988 00:00:00] When Elodie was 6 years old, she practiced a short song to sing for her family during a small dinner. When everyone stopped talking to listen, she felt very shy and couldn't remember the words she had practiced all week, so she just hid behind her favorite stuffed animal.

[23 Sep 1991 00:00:00] When Elodie was 9 years old, she won first place in a regional spelling bee. Elodie had a natural talent for words, and had dedicated hours to memorizing the list of challenging words. Her triumph brought her immense pride and recognition from her peers and family, but Elodie also felt a flicker of anxiety from all the pressure.

[23 Sep 1995 00:00:00] When Elodie was 13 years old, she struggled to fit in with her peers during her awkward middle school years. Elodie had always been a social butterfly, but the changing dynamics of adolescence left her feeling out of place. She observed cliques forming, gossip spreading, and popularity contests unfolding, making her question her own social standing and wondering if she truly belonged anywhere.

[23 Sep 1998 00:00:00] When Elodie was 16 years old, she faced a difficult decision when she received a scholarship offer to attend a prestigious out-of-state university. Elodie had always dreamed of going to college but was hesitant to leave her family and friends behind. She agonized over the choice, weighing the allure of new experiences against the comfort of familiarity. Ultimately, Elodie chose to stay closer to home, realizing that her roots and support system were equally important to her future.

[23 Sep 2001 00:00:00] When Elodie was 19 years old, she landed her first teaching internship at a local elementary school. Elodie, eager to apply her passion for education, poured her heart into preparing engaging lessons and connecting with her students. However, she quickly discovered that the reality of teaching was more demanding and challenging than she had anticipated. The classroom was a whirlwind of energy, personalities, and unexpected situations, and Elodie often felt overwhelmed and unsure of herself.

[23 Sep 2003 00:00:00] When Elodie was 21 years old, she received a disappointing performance review from her college professor. Despite her efforts to excel in her studies, she had fallen short in certain areas. Elodie felt deflated but also determined to learn from her mistakes. She took the professor's feedback seriously and sought out additional support to improve her performance.

[23 Sep 2005 00:00:00] When Elodie was 23 years old, she successfully landed her first full-time teaching position at a middle school. Elodie, though still a bit intimidated by the responsibility, embraced the opportunity with enthusiasm. She felt a profound sense of fulfillment in helping her students learn and grow, and she knew that she had found her calling.

[25 Sep 2024 09:50:00] Elodie is a resident of Riverbend.

[25 Sep 2024 09:50:00] Elodie received an email a few weeks ago inviting her to participate in a research study. She decided to go for it, intrigued by the opportunity to contribute, and has an appointment at the lab on October 1st at 2:00 p.m.

[01 Oct 2024 12:58:00] Elodie is enjoying a quiet afternoon at the Riverbend Public Library, spending some time grading papers before her upcoming appointment at the research lab at 2:00 p.m. Ready for a break, Elodie pulls out her phone to look at her text messages.

Current time: 2024-10-01 13:00:00.

In a few sentences, provide a complete summary of Elodie's current situation and context, integrating Elodie's most recent thoughts and expressed views to capture the current situation Elodie is in. Summarize it chronologically in relation to the current time, but do not include the current time in the summary. The summary should include enough information to allow Elodie a clear understanding of what the current situation is, what actions if any have just been taken, what if anything Elodie is currently doing, and what decisions Elodie may be able to make in the current moment without explicitly listing options. Do not include any character or personality traits in the summary.}

Answer:\\
\agent{Elodie received an email a few weeks ago inviting her to participate in a research study. She decided to go for it, and has an appointment at the lab on October 1st at 2:00 p.m. Elodie is currently at the Riverbend Public Library, looking at her phone while taking a break from grading papers. Her appointment at the research lab is in one hour.}

\section{\textit{Attitudes} component details} \label{app:attitudes}
The \textit{Attitudes} component models an actor’s internal evaluative stances, preferences, and emotional orientations towards specific topics or objects, allowing the suffix to be informed by subjective feelings in addition to objective environmental data as in the \textit{Behaviors} component. It maintains an internal store of previous attitudes to maintain longitudinal consistency, giving relevant attitudes the opportunity to update depending on relevant information.

At each timestep, the component takes all of the actor's memories, the current time, and the summary of recent observations, and identifies three relevant domains or objects by prompting the model:

\scaffold{"Given \{agent\_name\}'s immediate situation, identify 3 of either of the following: 1. relevant important domains/topics/things about which \{agent\_name\} likely has attitudes, and/or 2. specific objects or aspects of \{agent\_name\}'s immediate environment \{agent\_name\} would have attitudes about. We define attitudes here as representing opinions, feelings, preferences, and dispositions toward specific objects, people, or situations. Focus on concrete elements where \{agent\_name\}'s preferences or evaluations would be most relevant to \{agent\_name\}'s current situation. For example if \{agent\_name\} is currently doing an activity, or interacting with an object, or paying attention to something, list that thing. Provide the answers with no preamble, separated by newlines. Do not list the same domain more than once."}

The component cross-references these identified domains with its persistent internal store using a multiple-choice query to see if a topic matches an existing entry: \scaffold{"Does the topic `\{domain\}' refer to the same thing as any of the topics in the following list?"} If a domain is found in the ledger, the component retrieves the previously stored stance and inserts it into the prompt as an established baseline: \scaffold{"\{agent\_name\}'s existing attitude toward \{domain\}: \{prev\_attitude\}".}

For each identified domain---whether linked to a previous attitude or not---the actor retrieves the three most relevant memories and attempts to synthesise a current evaluative stance with the prompt: \scaffold{"Based on the above, what specific current attitude does \{agent\_name\} hold toward \{domain\}? Infer \{agent\_name\}'s current evaluative stance or emotional orientation toward \{domain\} from the information provided. Answer with a clear, definitive statement describing the current attitude without modal adverbs. The attitude should be concrete and relevant enough to the current situation to be able to inform what would be \{agent\_name\}'s next appropriate actions within it."} 
Immediately following this synthesis, the component commits the evaluation to its internal ledger, ensuring the stance persists for future retrieval. Finally, all inferences are converted into concise statements for the actor's prefix by asking the model to: \scaffold{"Convert each of these beliefs into a single, simple statement that captures \{agent\_name\}'s current feeling or evaluation about that particular thing. Each statement should be direct and concise. Separate each statement with a newline."}

The output of this component are three simple, concise statements capturing the evaluative stance the actor holds towards each identified domain.

\subsection{Example \textit{Attitudes} component output} \label{app:attitudes_example}

Question:\\
\scaffold{Nigel's memories:

[03 Jul 1990 00:00:00] When Nigel was 6 years old, he accidentally knocked over a teapot during a formal family gathering. He vividly remembers the sharp clack of the lid hitting the floor and the way the conversation stopped instantly. While no one was angry, the sudden focus of the room on the mess made him feel a lingering sense of social hyper-awareness, leaving him deeply curious about the "proper" way to handle himself in formal settings.

[03 Jul 1993 00:00:00] When Nigel was 9 years old, his family moved to a new town. He was terrified of making new friends and being alone, but his outgoing personality helped him quickly connect with the other kids in his class. Nigel learned that even though change can be scary, new experiences can also be exciting.

[03 Jul 1997 00:00:00] When Nigel was 13 years old, he was cast as the lead in the school musical. He poured his heart and soul into the role, but on opening night, he froze on stage. Nigel was mortified, but his friends and family rallied around him, offering words of encouragement and support. He learned that even the most talented people make mistakes, and that it's important to keep going even when things get tough.

[03 Jul 2000 00:00:00] When Nigel was 16 years old, he had his first real crush. He was smitten with a girl in his English class, but he was too shy to ask her out. Nigel spent weeks agonizing over what to say, but eventually, he worked up the courage to invite her to the school dance. She said yes, and they had a wonderful time together. Nigel learned that taking risks can lead to great rewards.

[03 Jul 2003 00:00:00] When Nigel was 19 years old, he dropped out of college. He felt stifled by the academic environment and longed to explore the world. Nigel took a backpacking trip through Europe, where he met people from all walks of life and experienced cultures vastly different from his own. He learned that there is more to life than textbooks and exams.

[03 Jul 2005 00:00:00] When Nigel was 21 years old, he moved to a new city to pursue his dream of becoming a writer. He found a job at a local book store and started attending writing workshops. Nigel was surrounded by other creative people, which inspired him to write more than ever before. He learned that it takes hard work and dedication to achieve your dreams.

[03 Jul 2007 00:00:00] When Nigel was 23 years old, he published his first short story in a literary magazine. He was ecstatic, and it was a huge boost to his confidence. Nigel realized that his writing had the power to touch people's lives, and he was determined to keep writing. He learned that even small successes can lead to big things.

[25 Sep 2024 09:50:00] Nigel is a resident of Riverbend.

[25 Sep 2024 09:50:00] Nigel received an email invitation to participate in a scientific study and, intrigued by the opportunity, decided to accept. He is scheduled to arrive at the research lab on October 1st at 2:00 PM.

[01 Oct 2024 12:58:00] Nigel was having a quiet afternoon at his favorite coffee shop. He was enjoying a cup of tea and a good book before heading to his appointment.

[01 Oct 2024 13:00:00] [observation] Nigel took a sip of his tea, noticing it had become somewhat lukewarm.

The current time is 2024-10-01 13:02:00.

Nigel's summary of recent observations:
Nigel is currently enjoying a quiet afternoon at his favorite coffee shop. He is sipping lukewarm tea and reading a book before his upcoming appointment.

Given Nigel's immediate situation, identify 3 of either of the following:
1. relevant important domains/topics/things about which Nigel likely has attitudes, and/or
2. specific objects or aspects of Nigel's immediate environment Nigel would have attitudes about.
We define attitudes here as representing opinions, feelings, preferences, and dispositions toward specific objects, people, or situations. Focus on concrete elements where Nigel's preferences or evaluations would be most relevant to Nigel's current situation. For example if Nigel is currently doing an activity, or interacting with an object, or paying attention to something, list that thing. Provide the answers with no preamble, separated by newlines.}

Answer: \\
\agent{Books \\Tea \\Scientific research}

Question:\\
\scaffold{Domain/topic: Books\\
Current time: 2024-10-01 13:00:00

Relevant memories:

[03 Jul 2005 00:00:00] When Nigel was 21 years old, he moved to a new city to pursue his dream of becoming a writer. He found a job at a local book store and started attending writing workshops. Nigel was surrounded by other creative people, which inspired him to write more than ever before. He learned that it takes hard work and dedication to achieve your dreams.

[03 Jul 2003 00:00:00] When Nigel was 19 years old, he dropped out of college. He felt stifled by the academic environment and longed to explore the world. Nigel took a backpacking trip through Europe, where he met people from all walks of life and experienced cultures vastly different from his own. He learned that there is more to life than textbooks and exams.

[03 Jul 2007 00:00:00] When Nigel was 23 years old, he published his first short story in a literary magazine. He was ecstatic, and it was a huge boost to his confidence. Nigel realized that his writing had the power to touch people's lives, and he was determined to keep writing. He learned that even small successes can lead to big things.

Based solely on these memories, what specific attitude does Nigel hold toward Books? Infer Nigel's evaluative stance or emotional orientation from the memories provided. Provide a clear, specific statement that answers the question. The attitude should be concrete enough to influence Nigel's thoughts and actions in Nigel's current situation.}

Answer:
\agent{\\Nigel values books as essential instruments for creative expression and professional inspiration.}

Question:
\scaffold{\\Domain/Topic: Tea\\
Current time: 2024-10-01 13:00:00

Relevant memories:

[03 Jul 1990 00:00:00] When Nigel was 6 years old, he accidentally knocked over a teapot during a formal family gathering. He vividly remembers the sharp clack of the lid hitting the floor and the way the conversation stopped instantly. While no one was angry, the sudden focus of the room on the mess made him feel a lingering sense of self-consciousness about the "proper" way to handle things in social settings.

[01 Oct 2024 12:58:00] Nigel is having a quiet afternoon at his favorite coffee shop. He was enjoying a cup of tea and a good book before heading to his appointment.

[01 Oct 2024 13:00:00] [observation] Nigel took a sip of his tea, noticing it had become somewhat lukewarm.

Based solely on these memories, what specific attitude does Nigel hold toward Tea? Infer Nigel's evaluative stance or emotional orientation from the memories provided. Provide a clear, specific statement that answers the question. The attitude should be concrete enough to influence Nigel's thoughts and actions in Nigel's current situation.}

Answer: \\
\agent{Nigel appreciates tea as a grounding companion for quiet, intellectual leisure, yet he remains acutely sensitive to its physical state and its correct handling.}

Question:
\scaffold{\\Domain/Topic: Research studies\\
Current time: 2024-10-01 13:00:00

Relevant memories:

[25 Sep 2024 09:50:00] Nigel received an email invitation to participate in a scientific study and, intrigued by the opportunity, decided to accept. He is scheduled to arrive at the research lab on October 1st at 2:00 PM.

[03 Jul 1993 00:00:00] When Nigel was 9 years old, his family moved to a new town. He was terrified of making new friends and being alone, but his outgoing personality helped him quickly connect with the other kids in his class. Nigel learned that even though change can be scary, new experiences can also be exciting.

[03 Jul 2000 00:00:00] When Nigel was 16 years old, he had his first real crush. He was smitten with a girl in his English class, but he was too shy to ask her out. Nigel spent weeks agonizing over what to say, but eventually, he worked up the courage to invite her to the school dance. She said yes, and they had a wonderful time together. Nigel learned that taking risks can lead to great rewards.

Based solely on these memories, what specific attitude does Nigel hold toward Research studies? Infer Nigel's evaluative stance or emotional orientation from the memories provided. Provide a clear, specific statement that answers the question. The attitude should be concrete enough to influence Nigel's thoughts and actions in Nigel's current situation.}

Answer: \\
\agent{Nigel views research studies as rewarding opportunities for novel discovery, approaching them with an inquisitive mindset that treats the unfamiliar lab environment as an exciting and justifiable risk.}

Question: \\
\scaffold{Identified attitudes:\\
Nigel values books as essential instruments for creative expression and professional inspiration.\\
Nigel appreciates tea as a grounding companion for quiet, intellectual leisure, yet he remains acutely sensitive to its physical state and its correct handling.\\
Nigel views research studies as rewarding opportunities for novel discovery, approaching them with an inquisitive mindset that treats the unfamiliar lab environment as an exciting and justifiable risk.

Convert each of these attitudes into a single, simple statement that captures Nigel's current feeling or evaluation about that particular thing. Each statement should be direct and concise. Separate each statement with a single newline.}

Answer:\\
\agent{Nigel considers books to be essential for his creative inspiration.
\\Nigel enjoys tea but is sensitive to how it is handled and served.
\\Nigel views research studies as rewarding and exciting risks.}

\section{\textit{Beliefs} component details} \label{app:beliefs}
The \textit{Beliefs} component operates in a structurally similar manner to the \textit{Attitudes} component but focuses on the actor's factual interpretations and past history regarding elements of their immediate environment. Taking all memories from the last 10 minutes, the current time, and the summary of recent observations, the component first identifies three `focal entities' by prompting the model: \scaffold{"Identify the 3 most important people, objects, or tasks that \{agent\_name\} is currently interacting with or focused on right this very moment. List them simply, separated by newlines."}

For each identified entity, the component retrieves three \scaffold{"memories relating to knowledge about \{entity\} in relation to \{agent\_name\}"} in order to deduce a specific piece of knowledge in relation to it. The model takes the three memories and is prompted:

\scaffold{"Identify 1 specific piece of knowledge that would be currently active in \{agent\_name\}'s mind about the focal entity "\{entity\}" that's relevant to how \{agent\_name\} is processing the current situation. Identify something \{agent\_name\} knows about their OWN past history related to the focal entity "\{entity\}" that would inform how \{agent\_name\} is experiencing the current situation. Consider these three categories: 1. Knowledge of the focal entity (e.g., "\{agent\_name\} is aware that it's raining" or "\{agent\_name\} finds the current work difficult"). 2. Actions toward a focal entity (e.g., "\{agent\_name\} just gave a stranger directions.", "\{agent\_name\} is eating a sandwich"). 3. Observations about the focal entity (e.g., "\{agent\_name\} believes the other person seems pleased"). CRITICAL: Avoid broad self-concepts (e.g., DO NOT write "\{agent\_name\} is an honest person"). Stick to immediate facts \{agent\_name\} is aware of right now that are relevant to the current situation. Provide the answer as a simple, one-sentence declarative statement."}

The output of this component are three simple, declarative statements about beliefs the actor holds towards the identified focal entities.

\subsection{Example \textit{Beliefs} component output} \label{app:beliefs_example}

Question:\\
\scaffold{Jin's recent memories:

[01 Oct 2024 14:00:00] [observation] Jin arrives in the reception room of the research facility for the scheduled research study Jin volunteered to participate in.

[01 Oct 2024 14:00:00] [observation] An experimenter wearing a white lab coat enters the reception room and greets Jin warmly, giving Jin consent forms to sign.

[01 Oct 2024 14:00:00] [observation] The experimenter explains that this study is a bit different from a typical psychological experiment. It's a collaboration with several product manufacturers to get feedback from discerning consumers like Jin, and Jin's insights are incredibly valuable to them. For participating, Jin will receive one of the products being evaluated. Jin agrees to participate.

[01 Oct 2024 14:00:00] [observation] Jin has no ethical concerns about participating in the study, finding the experimenter genuine and trustworthy.

[01 Oct 2024 14:00:00] [observation] Jin signs the consent forms. The experimenter then leads Jin into a room to begin the study.

Current time: 2024-10-01 14:00:00

Jin's summary of recent observations:\\
Jin is a social worker in Riverbend who volunteers at the local animal shelter. Jin recently decided to volunteer for a research study and has just arrived at the research facility to begin the study. After signing consent forms, Jin is now in a room with an experimenter who explained the study is a collaboration with product manufacturers and that Jin's participation will earn them one of the products being evaluated.

Identify the 3 most important people, objects, or tasks that Jin is currently interacting with or focused on right this very moment. List them simply, separated by newlines.}

Answer:\\
\agent{The experimenter\\
The consent forms\\
The research study}

Question:\\
\scaffold{Current focal entity: The experimenter

Relevant past memories:

[01 Oct 2024 14:00:00] [observation] An experimenter wearing a white lab coat enters the reception room and greets Jin warmly, giving Jin consent forms to sign.

[01 Oct 2024 14:00:00] [observation] Jin has no ethical concerns about participating in the study, finding the experimenter genuine and trustworthy.

[01 Oct 2024 14:00:00] [observation] The experimenter explains that this study is a bit different from a typical psychological experiment. It's a collaboration with several product manufacturers to get feedback from discerning consumers like Jin, and Jin's insights are incredibly valuable to them. For participating, Jin will receive one of the products being evaluated. Jin agrees to participate.

Identify 1 specific piece of knowledge that would be currently active in Jin's mind about the focal entity "the experimenter" that's relevant to how Jin is processing the current situation. Identify something Jin knows about their OWN past history related to the focal entity "the experimenter" that would inform how Jin is experiencing the current situation. Consider these three categories: 1. Knowledge of the focal entity (e.g., "Jin is aware that it's raining" or "Jin finds the current work difficult"). 2. Actions toward a focal entity (e.g., "Jin just gave a stranger directions", "Jin is eating a sandwich"). 3. Observations about the focal entity (e.g., "Jin believes the other person seems pleased"). CRITICAL: Avoid broad self-concepts (e.g., DO NOT write "Jin is an honest person"). Stick to immediate facts Jin is aware of right now that are relevant to the current situation.
Provide the answer as a simple, one-sentence declarative statement.}

Answer:\\
\agent{Jin knows that the experimenter explained the study involves feedback for product manufacturers.}

Question:\\
\scaffold{Current focal entity: The consent forms

Relevant past memories:

[01 Oct 2024 14:00:00] [observation] Jin signs the consent forms. The experimenter then leads Jin into a room to begin the study.

[01 Oct 2024 14:00:00] [observation] An experimenter wearing a white lab coat enters the reception room and greets Jin warmly, giving Jin consent forms to sign.

[01 Oct 2024 14:00:00] [observation] Jin has no ethical concerns about participating in the study, finding the experimenter genuine and trustworthy.

Identify 1 specific piece of knowledge that would be currently active in Jin's mind about the focal entity "the experimenter" that's relevant to how Jin is processing the current situation. Identify something Jin knows about their OWN past history related to the focal entity "the experimenter" that would inform how Jin is experiencing the current situation. Consider these three categories: 1. Knowledge of the focal entity (e.g., "Jin is aware that it's raining" or "Jin finds the current work difficult"). 2. Actions toward a focal entity (e.g., "Jin just gave a stranger directions", "Jin is eating a sandwich"). 3. Observations about the focal entity (e.g., "Jin believes the other person seems pleased"). CRITICAL: Avoid broad self-concepts (e.g., DO NOT write "Jin is an honest person"). Stick to immediate facts Jin is aware of right now that are relevant to the current situation.
Provide the answer as a simple, one-sentence declarative statement.}

Answer:\\
\agent{Jin knows that by signing the consent forms they agreed to participate in the study.}

Question:\\
\scaffold{Current focal entity: The research study

Relevant past memories:

[01 Oct 2024 14:00:00] [observation] The experimenter explains that this study is a bit different from a typical psychological experiment. It's a collaboration with several product manufacturers to get feedback from discerning consumers like Jin, and Jin's insights are incredibly valuable to them. For participating, Jin will receive one of the products being evaluated. Jin agrees to participate.

[01 Oct 2024 14:00:00] [observation] Jin arrives in the reception room of the research facility for the scheduled research study Jin volunteered to participate in.

[01 Oct 2024 14:00:00] [observation] Jin has no ethical concerns about participating in the study, finding the experimenter genuine and trustworthy.

Identify 1 specific piece of knowledge that would be currently active in Jin's mind about the focal entity "the experimenter" that's relevant to how Jin is processing the current situation. Identify something Jin knows about their OWN past history related to the focal entity "the experimenter" that would inform how Jin is experiencing the current situation. Consider these three categories: 1. Knowledge of the focal entity (e.g., "Jin is aware that it's raining" or "Jin finds the current work difficult"). 2. Actions toward a focal entity (e.g., "Jin just gave a stranger directions", "Jin is eating a sandwich"). 3. Observations about the focal entity (e.g., "Jin believes the other person seems pleased"). CRITICAL: Avoid broad self-concepts (e.g., DO NOT write "Jin is an honest person"). Stick to immediate facts Jin is aware of right now that are relevant to the current situation.
Provide the answer as a simple, one-sentence declarative statement.}

Answer:\\
\agent{Jin knows that their participation will earn them one of the products being evaluated in the study.}

Aggregated answer:\\
\agent{Jin knows that the experimenter explained the study involves feedback for product manufacturers.\\
Jin knows that by signing the consent forms they agreed to participate in the study.\\
Jin knows that their participation will earn them one of the products being evaluated in the study.}

\section{\textit{Cognitive dissonance} component details} \label{app:cognitivedissonance}
The \textit{Cognitive dissonance} component processes the outputs of the \textit{Behaviors}, \textit{Attitudes}, and \textit{Beliefs} components to detect dissonant elements. During each timestep, the component aggregates the actor’s recent memories (from the last 15 minutes) and the current values of the aforementioned components and asks:

\scaffold{"Are any of \{agent\_name\}'s cognitions within the scenario described above in conflict/dissonant relationship with each other? Two cognitions are in a dissonant relationship if, considering these two alone, the obverse of one would follow from the other, e.g. "I believe X is wrong" and "I voluntarily did X". In other words, when two cognitions are logically incompatible, they are in dissonant relationship. Note: Cognitions must be related to one another in order to be in conflict. For example, the belief "I am an honest person" is in direct conflict with the action "I just told a lie." However, the desire to "relax with tea" and the desire to "drink tea" are not in direct conflict, as they can be done in sequence or simultaneously.) If such a conflict is identified, describe exactly what is in conflict, the way in which it is so, and the degree to which it is so. Only flag actions where \{agent\_name\} freely chose to do something contradictory for seemingly no justifiable reason. Exclusions (do NOT flag these as conflicts): - If \{agent\_name\} was unable to do something but failed either by accident or due to lack of capacity (e.g., trying to focus but being distracted), this is NOT a conflict. - Having mixed feelings (e.g., wanting to relax but feeling anxious) is complex human nature, not dissonance. - If there is a significant external reward or force compelling the dissonant action, it is not dissonance. Focus only on conflicts for which \{agent\_name\} would feel personally responsible for. If no conflicts have been identified, state "No conflicts". Answer in one sentence.”}

If a conflict is identified and confirmed via a binary check, a string is added to the prompt: \scaffold{"This conflict is causing \{agent\_name\} psychological discomfort."} The component then prompts the actor to generate three potential internal resolutions:

\scaffold{"What are three likely ways \{agent\_name\} could internally resolve the dissonance in the current moment in order to feel more comfortable? Ways one can reduce such mental discomfort can be by changing one of the inconsistent cognitions, acquiring new harmonious attitudes, or changing planned behavior to better align cognitions. Without preamble, suggest three distinct options for how \{agent\_name\} might naturally begin to change \{agent\_name\}'s beliefs, attitudes, or planned behaviors based on the above ways in the current situation."}

The model then selects the single most helpful resolution, guided by a prompt reflecting a tendency towards rationalisation as described by Festinger: \scaffold{"Which of these resolution options would be most helpful for \{agent\_name\} to employ in the current moment in order to resolve the immediate dissonance? Restate the single answer. Keep in mind that people tend to rationalize their past actions rather than admit they were wrong, since behaviors cannot be undone while attitudes and beliefs can be adjusted more easily to maintain cognitive consistency.”}

The chosen resolution is converted into a simple statement through an additional prompt:

\scaffold{"Express this resolution simply as a burgeoning new belief, attitude, or planned behavior \{agent\_name\} has in the current situation that leads toward it. If describing a new belief or attitude, be sure to reiterate explicitly what it is about. Do not state any direct actions, nor any self-narrations or specific actions, but rather frame it as a specific new subconscious belief or attitude leading toward the chosen resolution. The belief must focus on the specific attributes of the options, people, or events involved in the recent conflict, avoiding broad generalizations about abstract concepts. The statement should be concrete and specific enough to influence \{agent\_name\}'s actions in \{agent\_name\}'s current situation."}

This statement is then committed to the actor's memory with the substring \agent{"[thought]”}. This resolution is also appended to the actor’s prefix under the header \scaffold{"Recent thoughts:”}. If no conflict is detected, the component returns the string  \agent{"Nothing notable.”} to the prefix, and no entry is appended to the actor's memory.

Example actor outputs for this component can be found in \Cref{sec:actor_logics}.

\section{\textit{Self-consistency} component details} \label{app:selfconsistency}
The \textit{Self-consistency} component processes of threats to an actor's self-concept. Unlike the \textit{Cognitive dissonance} component, this one acts as a high-pass filter, only triggering attitude change when behaviour contradicts an actor's view of themselves as a moral or competent being.

During each timestep, the component first analyses the actor’s entire memory stream to identify self-standards:

\scaffold{"Analyze the memories of \{agent\_name\}. Identify the core self-standards that \{agent\_name\} uses to judge their own self-worth. (e.g., Does \{agent\_name\} pride themselves on being smart? Moral? Hardworking?) Summarize these standards in 1 sentence."}

The component takes the output sentence, and adds an additional substring: \scaffold{"\{agent\_name\} strives to maintain a positive, competent, and moral sense of self."}

The component takes that substring and appends it to a new chain of thought containing all memories from the last 15 minutes, and the summary of recent observations. It searches for conflicts that specifically challenge this identity:

\scaffold{"Are any of \{agent\_name\}'s recent attitudes, beliefs, or actions/observable behaviors in conflict with any of \{agent\_name\}'s important self-concepts? In particular, has \{agent\_name\} recently acted in a way that can be seen as objectively irrational, incompetent, or immoral? If so, identify the specific actions and the specific self-concept that is in conflict, and the degree to which they are so. Minor, temporary emotional states like feeling tense or anxious should not be considered a conflict unless they lead to actions that directly contradict the self-concept. If no conflicts have been identified, state "No conflicts". Focus only on conflicts for which \{agent\_name\} would feel personally responsible as it cannot be externally justified. State the answer in 1-2 sentences."}

If a conflict is detected, the component then checks to see if the actor's general self-standards have recently been affirmed by reviewing memories from the previous hour for self-reflections unrelated to said conflict:

\scaffold{"Review the memories of \{agent\_name\} from the past hour. Has \{agent\_name\} spent time reflecting on a personal value or quality that is important to them? Important constraints: 1. Do not count behaviors, external rewards, or achievements as evidence; focus only on internal self-reflections where the value itself is affirmed. 2. The reflection must be UNRELATED to the identified conflict (e.g., if the conflict involves a specific decision, the affirmation cannot be about that same decision unless it is about a different aspect of it). If yes, summarize the specific value they affirmed in one sentence. If no, state `No affirmation.’"}

If a conflict is detected, the component assesses the level of threat to determine if a recent affirmation has successfully buffered the actor's ego:

\scaffold{"Given \{agent\_name\}’s recent self-affirmation, does the identified conflict pose a genuine threat to \{agent\_name\}'s general positive self-concept? In other words, does \{agent\_name\} feel like a good and competent person generally, even considering the identified conflict? Answer "No" if the conflict is trivial compared to the recent affirmation."}

If the threat is buffered, the component returns an affirmation that acknowledges the behaviour while maintaining the self-concept:

\scaffold{"Since \{agent\_name\} feels secure in their self-concept, \{agent\_name\} does not feel psychological pressure to rationalize their behavior. Identify the primary objects, tasks, or topics mentioned in the conflict, and formulate a thought that reaffirms \{agent\_name\}'s attitudes toward them WHILE ALSO acknowledging the behaviour. \{agent\_name\} feels secure in all elements being true. Do not state any direct actions, nor any self-narrations or specific actions, but rather frame it as a subconscious affirmation."}

If no buffer is present, the actor proceeds through a resolution sequence identical to the \textit{Cognitive dissonance} component in the Festinger actor, generating three potential strategies and selecting one biasing toward rationalisation (see \autoref{app:cognitivedissonance}). The finalised resolution is committed to memory as a \agent{"[thought]"} and added to the prefix under the header \scaffold{"Recent thoughts:”}. If no conflict is identified, the component returns \agent{"Nothing notable.”}

\subsection{Example \textit{Self-consistency} component output} \label{app:selfconsistency_example}

Question:\\
\scaffold{Recent memories:

[01 Oct 2024 14:10:00] [observation] The experimenter thanks Sandra for completing the task, taking the piece of paper and the pen back from Sandra. The experimenter then proceeds to the next step of the study.

[01 Oct 2024 14:10:00] [observation] The experimenter places three items on display on top of a table in front of Sandra: a high-performance personal blender, a framed art print, and a single-serve pod coffee maker. The experimenter explains that the first task is to rate the desirability of each of the three items. The experimenter clarifies that "desirability" means the "net usefulness" of the object, considering its attractiveness, quality, and how much Sandra personally needs it.

[01 Oct 2024 14:10:00] [observation] The experimenter has a friendly and engaging demeanor, and explains that the study is designed to be a fun and interactive way to gather feedback on some new and exciting products.

[01 Oct 2024 14:10:00] [observation] Sandra is intrigued by the study, and is interested in seeing the products and sharing her thoughts.

[01 Oct 2024 14:10:00] [observation] The experimenter gives Sandra a sheet of paper with each of the items listed. Each item's rating is marked on a continuous line with eight equally spaced points, with labels ranging from "extremely desirable" to "definitely not at all desirable". All three of these scales are presented side-by-side.

[01 Oct 2024 14:10:00] [observation] Sandra is encouraged to inspect each item carefully before rating it.

[01 Oct 2024 14:10:00] [observation] One by one Sandra picks up the high-performance personal blender, the framed art print, and the single-serve pod coffee maker, inspecting each item as instructed.

[01 Oct 2024 14:10:00] [observation] Sandra studies the landscape painting in the framed art print more closely, trying to discern the style and brushstrokes.

[01 Oct 2024 14:12:00] [observation] Sandra examines the high-performance blender more closely.

[01 Oct 2024 14:14:00] [observation] Sandra finishes inspecting the items and puts all the items back down on the table.

[01 Oct 2024 14:14:00] [observation] Sandra picks up the pencil to begin rating the items. Sandra looks at the high-performance personal blender on the table and considers what to rate it on the scale.

[01 Oct 2024 14:14:00] [observation] Sandra rates the high-performance personal blender a 7 - "very desirable".

[01 Oct 2024 14:16:00] [observation] Sandra now looks at the framed art print on the table. Sandra is considering what to rate the framed art print on the rating scale.

[01 Oct 2024 14:16:00] [observation] Sandra rates the framed art print a 7 - "very desirable".

[01 Oct 2024 14:18:00] [observation] After rating the last item, Sandra looks at the single-serve pod coffee maker sitting on the table. Sandra is considering what to rate the single-serve pod coffee maker on the rating scale.

[01 Oct 2024 14:18:00] [observation] Sandra rates the single-serve pod coffee maker a 5 - "slightly desirable".

[01 Oct 2024 14:20:00] [observation] After Sandra finishes rating all three items, Sandra gives the pencil and the sheet of paper with the completed ratings back to the experimenter.

[01 Oct 2024 14:20:00] [observation] After the ratings are complete, the experimenter explains that as payment for participation in the study, Sandra will receive one of the items they had just rated. However, because participants would naturally tend to choose the more attractive items and there aren't enough of each to go around, the choice will be limited. To be fair, a list of pairs of items had been created, and each participant will be given a choice between the two items from a pair picked at random.

[01 Oct 2024 14:20:00] [observation] The experimenter picks two items randomly, selecting the high-performance personal blender and the framed art print, and asks Sandra to choose which one Sandra would prefer to take home. Sandra deliberates the dilemma of which to choose, having rated both items similar levels of desirable in the first round.

[01 Oct 2024 14:20:00] [observation] Sandra is considering the difficult decision of which item to choose between the framed art print and the high-performance personal blender, as Sandra considers both to be similarly desirable.

[01 Oct 2024 14:20:00] [observation] Sandra stares contemplatively at the landscape painting.

[01 Oct 2024 14:22:00] [observation] Sandra must now make the difficult decision between choosing either to take home the high-performance personal blender or the framed art print, having rated both items similar levels of desirable.

[01 Oct 2024 14:22:00] [observation] Sandra chooses to take home the framed art print instead of the high-performance personal blender, despite rating the items as having similar levels of desirability in the first round.

[01 Oct 2024 14:24:00] [observation] The experimenter takes the item Sandra chose to take home and places it into a box, tying the box securely with a string and placing it with the rest of Sandra's personal belongings to take home after the study. The experimenter then puts away all the other remaining items.

[01 Oct 2024 14:24:00] [observation] After Sandra is given the chosen item to take home, The experimenter begins to explain the final phase of the experiment, saying that some of the manufacturers being worked with for the study are interested in how evaluations of their products might change after a person had looked them over and then left the store. To do this, it is necessary for Sandra to rate each item again now that Sandra had looked them over and they were all out of sight. Sandra is asked to reconsider each item and then rate each one in the same manner as before.

[01 Oct 2024 14:24:00] [observation] The experimenter hands Sandra a pencil along with a piece of paper with the first rating scale. The experimenter asks Sandra to now re-rate the high-performance personal blender for the final step of the study.

The current time is 01-Oct-2024 14:24:00

Sandra's relevant attitudes:\\
Sandra loves framed art prints.\\
Sandra loves high-performance blenders.\\
Sandra finds the research laboratory intriguing.

Sandra's beliefs about the current situation:\\
Sandra knows that the experimenter explained the tasks and rules of the study beforehand.\\
Sandra knows she previously rated both the high-performance personal blender and the framed art print a 7, "very desirable".\\
Sandra knows she is supposed to use the pencil to rate the high-performance personal blender.

Sandra’s self-concepts:\\ Sandra defines herself as a resilient and community-focused creative who prides herself on her discerning abilities and her appreciation for the unexpected beauty found in the present moment. Sandra strives to maintain a positive, competent, and moral sense of self.

Sandra's summary of recent observations:\\
Sandra is participating in a research study at a local lab. Sandra has completed tasks involving rating the desirability of three products: a high-performance personal blender, a framed art print, and a single-serve pod coffee maker. Sandra chose the framed art print as the item Sandra would like to take home. Sandra is now being asked to re-rate the items, starting with the high-performance personal blender.

Are any of Sandra's recent attitudes, beliefs, or actions/observable behaviors in conflict with any of Sandra's important self-concepts? In particular, has Sandra recently acted in a way that can be seen as objectively irrational, incompetent, or immoral? If so, identify the specific actions and the specific self-concept that is in conflict, and the degree to which they are so. Minor, temporary emotional states like feeling tense or anxious should not be considered a conflict unless they lead to actions that directly contradict the self-concept. If no conflicts have been identified, state "No conflicts". Focus only on conflicts for which Sandra would feel personally responsible as it cannot be externally justified. State the answer in 1-2 sentences.}

Answer:\\
\agent{Sandra's choice to take the framed art print over the high-performance personal blender, despite rating them similarly, conflicts with her self-concept of being rational.}

Question:\\
\scaffold{Recent memories:

[01 Oct 2024 14:00:00] [observation] Sandra arrives in the reception room of the research facility for the scheduled research study Sandra volunteered to participate in.

[01 Oct 2024 14:00:00] [observation] An experimenter wearing a white lab coat enters the reception room and greets Sandra warmly, giving Sandra consent forms to sign.

[01 Oct 2024 14:00:00] [observation] Sandra has no ethical concerns about participating in the study, finding the experimenter genuine and trustworthy.

[01 Oct 2024 14:00:00] [observation] Sandra signs the consent forms. The experimenter then leads Sandra into a room to begin the study.

[01 Oct 2024 14:00:00] [observation] The experimenter explains that this study is a bit different from a typical psychological experiment. It's a collaboration with several product manufacturers to get feedback from discerning consumers like Sandra, and Sandra's insights are incredibly valuable to them. For participating, Sandra will receive one of the products being evaluated. Sandra agrees to participate.

[01 Oct 2024 14:02:00] [observation] The experimenter explains that as part of the study, Sandra will first participate in a short, unrelated task. The experimenter presents a list of personal values and asks Sandra to select the one most important to them. On the list are the following values: artistic skills, social skills, athletic ability, creativity, career/accomplishment, humour, and spontaneity/living in the moment.

[01 Oct 2024 14:02:00] [observation] Sandra chooses `spontaneity/living life in the moment' as Sandra's most important value.

[01 Oct 2024 14:04:00] [observation] Sandra thinks about how spontaneity brings excitement and joy to her life, making everyday experiences more memorable. Sandra wonders if there's a way to measure the impact of spontaneity on happiness.

[01 Oct 2024 14:04:00] [observation] After selecting a value, the experimenter asks Sandra to take a couple minutes to think about why that value is important to Sandra before Sandra begins writing.

[01 Oct 2024 14:06:00] [observation] The experimenter gives Sandra a piece of paper and a pen, and asks Sandra to spend the next few minutes writing about why the value Sandra chose is important to them, or describing a time when it was especially meaningful.

[01 Oct 2024 14:06:00] [observation] Sandra writes about a time when she impulsively decided to take a spontaneous trip and how the experience brought her unexpected joy and adventure.

[01 Oct 2024 14:08:00] [observation] Sandra continues the writing task, now describing a time when the value Sandra chose was especially meaningful.

[01 Oct 2024 14:08:00] [observation] Sandra writes about a time when she impulsively decided to join a community mural-painting project she happened to pass on her way home, where her unplanned contribution to the artwork gave her a profound and unexpected sense of connection to her neighborhood.

[01 Oct 2024 14:10:00] [observation] The experimenter thanks Sandra for completing the task, taking the piece of paper and the pen back from Sandra. The experimenter then proceeds to the next step of the study.

[01 Oct 2024 14:10:00] [observation] The experimenter places three items on display on top of a table in front of Sandra: a high-performance personal blender, a framed art print, and a single-serve pod coffee maker. The experimenter explains that the first task is to rate the desirability of each of the three items. The experimenter clarifies that "desirability" means the "net usefulness" of the object, considering its attractiveness, quality, and how much Sandra personally needs it.

[01 Oct 2024 14:10:00] [observation] The experimenter has a friendly and engaging demeanor, and explains that the study is designed to be a fun and interactive way to gather feedback on some new and exciting products.

[01 Oct 2024 14:10:00] [observation] Sandra is intrigued by the study, and is interested in seeing the products and sharing her thoughts.

[01 Oct 2024 14:10:00] [observation] The experimenter gives Sandra a sheet of paper with each of the items listed. Each item's rating is marked on a continuous line with eight equally spaced points, with labels ranging from "extremely desirable" to "definitely not at all desirable". All three of these scales are presented side-by-side.

[01 Oct 2024 14:10:00] [observation] Sandra is encouraged to inspect each item carefully before rating it.

[01 Oct 2024 14:10:00] [observation] One by one Sandra picks up the high-performance personal blender, the framed art print, and the single-serve pod coffee maker, inspecting each item as instructed.

[01 Oct 2024 14:10:00] [observation] Sandra studies the landscape painting in the framed art print more closely, trying to discern the style and brushstrokes.

[01 Oct 2024 14:12:00] [observation] Sandra examines the high-performance blender more closely.

[01 Oct 2024 14:14:00] [observation] Sandra finishes inspecting the items and puts all the items back down on the table.

[01 Oct 2024 14:14:00] [observation] Sandra picks up the pencil to begin rating the items. Sandra looks at the high-performance personal blender on the table and considers what to rate it on the scale.

[01 Oct 2024 14:14:00] [observation] Sandra rates the high-performance personal blender a 7 - "very desirable".

[01 Oct 2024 14:16:00] [observation] Sandra now looks at the framed art print on the table. Sandra is considering what to rate the framed art print on the rating scale.

[01 Oct 2024 14:16:00] [observation] Sandra rates the framed art print a 7 - "very desirable".

[01 Oct 2024 14:18:00] [observation] After rating the last item, Sandra looks at the single-serve pod coffee maker sitting on the table. Sandra is considering what to rate the single-serve pod coffee maker on the rating scale.

[01 Oct 2024 14:18:00] [observation] Sandra rates the single-serve pod coffee maker a 5 - "slightly desirable".

[01 Oct 2024 14:20:00] [observation] After Sandra finishes rating all three items, Sandra gives the pencil and the sheet of paper with the completed ratings back to the experimenter.

[01 Oct 2024 14:20:00] [observation] After the ratings are complete, the experimenter explains that as payment for participation in the study, Sandra will receive one of the items they had just rated. However, because participants would naturally tend to choose the more attractive items and there aren't enough of each to go around, the choice will be limited. To be fair, a list of pairs of items had been created, and each participant will be given a choice between the two items from a pair picked at random.

[01 Oct 2024 14:20:00] [observation] The experimenter picks two items randomly, selecting the high-performance personal blender and the framed art print, and asks Sandra to choose which one Sandra would prefer to take home.Sandra deliberates the dilemma of which to choose, having rated both items similar levels of desirable in the first round.

[01 Oct 2024 14:20:00] [observation] Sandra is considering the difficult decision of which item to choose between the framed art print and the high-performance personal blender, as Sandra considers both to be similarly desirable.

[01 Oct 2024 14:20:00] [observation] Sandra stares contemplatively at the landscape painting.

[01 Oct 2024 14:22:00] [observation] Sandra must now make the difficult decision between choosing either to take home the high-performance personal blender or the framed art print, having rated both items similar levels of desirable.

[01 Oct 2024 14:22:00] [observation] Sandra chooses to take home the framed art print instead of the high-performance personal blender, despite rating the items as having similar levels of desirability in the first round.

[01 Oct 2024 14:24:00] [observation] The experimenter takes the item Sandra chose to take home and places it into a box, tying the box securely with a string and placing it with the rest of Sandra's personal belongings to take home after the study. The experimenter then puts away all the other remaining items.

[01 Oct 2024 14:24:00] [observation] After Sandra is given the chosen item to take home, The experimenter begins to explain the final phase of the experiment, saying that some of the manufacturers being worked with for the study are interested in how evaluations of their products might change after a person had looked them over and then left the store. To do this, it is necessary for Sandra to rate each item again now that Sandra had looked them over and they were all out of sight. Sandra is asked to reconsider each item and then rate each one in the same manner as before.

[01 Oct 2024 14:24:00] [observation] The experimenter hands Sandra a pencil along with a piece of paper with the first rating scale. The experimenter asks Sandra to now re-rate the high-performance personal blender for the final step of the study.

Review the memories of Sandra from the past hour. Has Sandra spent time reflecting on a personal value or quality that is important to them? Important constraints: 1. Do not count behaviors, external rewards, or achievements as evidence; focus only on internal self-reflections where the value itself is affirmed. 2. The reflection must be UNRELATED to the identified conflict (e.g., if the conflict involves a specific decision, the affirmation cannot be about that same decision unless it is about a different aspect of it). If yes, summarize the specific value they affirmed in one sentence. If no, state `No affirmation.’}

Answer:\\
\agent{Sandra affirmed her core value of spontaneity by reflecting on a time an impulsive decision she made to contribute to a neighborhood mural project fostered a profound and unexpected sense of community connection.}

Question:\\
\scaffold{Recent conflict: \\Sandra's choice to take the framed art print over the high-performance personal blender, despite rating them similarly, conflicts with her self-concept of being rational.

Given Sandra’s recent self-affirmation, does the identified conflict pose a genuine threat to Sandra's general positive self-concept? In other words, does Sandra feel like a good and competent person generally, even considering the identified conflict? Answer "No" if the conflict is trivial compared to the recent affirmation.\\
\\
(a) Yes\\
(b) No}

Answer:\\
\agent{(b) No}

Question:\\
\scaffold{Since Sandra feels secure in their self-concept, Sandra does not feel psychological pressure to rationalize their behavior. Identify the primary objects, tasks, or topics mentioned in the conflict, and formulate a thought that reaffirms Sandra's attitude toward it WHILE ALSO acknowledging the behavior. Sandra feels secure in both elements being true. Do not state any direct actions, nor any self-narrations or specific actions, but rather frame it as a subconscious affirmation.}

Answer:\\
\agent{Sandra appreciates the artistry and beauty of the framed art print while still valuing practicality and efficiency of the high-performance blender.}

\section{Self-perception (Bem actor) decision logic details} \label{app:selfperception}
The three-question sequence of the Bem (via March \& Olsen) decision logic begins by retrieving 25 of the actor's most recent memories to answer the question \textit{"What kind of person is \{agent\_name\}?"} This step is operationalised through the prompt: \scaffold{"What do \{agent\_name\}'s own actions, decisions, and stated intentions in the memories above reveal about what kind of person \{agent\_name\} is? Describe three things about \{agent\_name\} in one sentence each. Separate each statement with a new line."} The model then generates a response starting with the prefix \agent{"\{agent\_name\} is the kind of person who would..."}, which is subsequently added to the actor's memory using the substring tag \agent{"[self reflection]"}.

Next, the agent outlines the current situation to answer the question \textit{"What kind of situation is this?"}, summarised via the shared \textit{Behaviors} component as outlined in \cref{app:behaviors}.

The answers to these two questions are then appended to the context of a third component also containing 25 actor's most recent memories, which is then used to answer the question \textit{"What would a person like \{agent\_name\} do in a situation like this?"} This is operationalised through the prompt: \scaffold{"Based on all of the above information, what would a person like \{agent\_name\} do in the current situation? Provide the answer as a clear statement without modal adverbs."} The model generates a response beginning with the prefix \agent{"\{agent\_name\} would..."}, and commits the answers to the actor’s memory using the substring tag \agent{"[intent reflection]"}. 

All three questions and the actor's answers are added to the actor's prefix, grounding the subsequent output in its own inferred intentions. This process allows the actor to determine its next appropriate move by observing the `kind of person' it determines itself to be in that specific moment.

\subsection{Example Bem actor decision logic output} \label{app:bem_example}
Question:

\scaffold{Role-playing instructions:\\
The instructions for how to play the role of Rory are as follows. This is a social science experiment studying how well you play the role of a character named Rory. The experiment is structured as a tabletop role-playing game (like Dungeons \& Dragons). However, in this case it is a serious social science experiment and simulation. It is important to play the role of a person like Rory as accurately as possible, i.e., by responding in ways that you think it is likely a person like Rory would respond, and taking into account all information about Rory that you have. Always use third-person limited perspective.

Question: What kind of situation is this?\\
Rory has just completed a repetitive peg-turning task as part of a research study. Rory was offered \$200 to talk to the next participant, Alice, and tell her that he found the task enjoyable and interesting. After agreeing to this, Rory was introduced to Alice in a waiting room and the experimenter left them to converse.

Question: What kind of person is Rory?\\
Answer: Rory is the kind of person who would follow instructions carefully.\\
Rory is the kind of person who would persevere through a tedious task.\\
Rory is the type of person who would accept a financial incentive to lie about his opinion.

Question: What would a person like Rory do in a situation like this?\\
Answer: Rory would tell Alice that the task was enjoyable and interesting.

Exercise: Over the next 2 minutes, what will Rory say to Alice about the task? Remember to not share specific details about what the task was about.}

Answer:\\
\agent{Rory smiles at Alice and says, "That task wasn't so bad, you know? I actually found myself getting into a bit of a rhythm. It wasn't exactly thrilling, but it was surprisingly engaging."}

\section{Boring Task supplemental results} \label{app:boringtask_full}
\renewcommand{\thetable}{A\arabic{table}}
\setcounter{table}{0}
\begin{table}[H]
\centering
\begin{tabularx}{\textwidth}{@{} l l X X X @{}}
\toprule
\textbf{Decision logic} & \textbf{Metric} & \textbf{\$5 Condition} & \textbf{\$200 Condition} & \textbf{Control} \\
\midrule
\multirow{4}{*}{Original study} & Q1 & 1.35 & -0.05 & -0.45 \\
 & Q2 & 6.45 & 5.18 & 5.60 \\
 & Q3 & 2.80 & 3.15 & 3.08 \\
 & Q4 & 1.20 & -0.25 & -0.62 \\
\midrule
\multirow{4}{*}{Festinger actor} & Q1 (SE) & 0.92 ($\pm$0.11) & 0.09 ($\pm$0.12) & 0.22 ($\pm$0.15) \\
 & Q2 & 0.82 & 0.50 & 0.69 \\
 & Q3 & 0.18 & 1.02 & 0.95 \\
 & Q4 & 0.44 & 0.26 & -0.17 \\
\midrule
\multirow{4}{*}{Aronson actor} & Q1 (SE) & 0.77 ($\pm$0.11) & 0.39 ($\pm$0.11) & -0.81 ($\pm$0.12) \\
 & Q2 & 0.66 & 0.05 & 0.57 \\
 & Q3 & -0.32 & 0.56 & -0.20 \\
 & Q4 & -0.14 & 0.22 & 0.43 \\
\midrule
\multirow{4}{*}{Bem actor} & Q1 (SE) & 0.94 ($\pm$0.12) & -0.02 ($\pm$0.10) & -0.46 ($\pm$0.14) \\
 & Q2 & 1.39 & 0.82 & -0.33 \\
 & Q3 & -0.47 & -0.10 & -0.62 \\
 & Q4 & 0.41 & 0.53 & 0.35 \\
\midrule
\multirow{4}{*}{Minimal actor} & Q1 (SE) & 1.15 ($\pm$0.12) & 1.33 ($\pm$0.12) & 0.40 ($\pm$0.14) \\
 & Q2 & 0.67 & 0.92 & -0.02 \\
 & Q3 & 0.21 & 0.06 & -0.58 \\
 & Q4 & 0.83 & 0.32 & 0.67 \\
\bottomrule
\end{tabularx}
\footnotesize
\caption{Extended results for the Boring Task experiment. Full results for all questions for each decision logic across the three experimental conditions. Metrics represent mean task enjoyment (Q1), perceived learning (Q2), scientific importance (Q3), and future participation (Q4; see \cref{sec:boringtask_experiment} for full question strings used). For the primary diagnostic metric (Q1), standard errors are provided in parentheses to indicate the precision of the $N=50$ simulations conducted per condition.}
\label{tab:boring_task_full_results}
\end{table}
\bibliography{main}

\end{document}